%% file: sn-article.tex
\documentclass[sn-mathphys,Numbered]{sn-jnl}

\usepackage{graphicx}%
\usepackage{multirow}%
\usepackage{amsmath,amssymb,amsfonts}%
\usepackage{amsthm}%
\usepackage{mathrsfs}%
\usepackage[title]{appendix}%
\usepackage{xcolor}%
\usepackage{textcomp}%
\usepackage{manyfoot}%
\usepackage{booktabs}%
\usepackage{algorithm}%
\usepackage{algorithmicx}%
\usepackage{algpseudocode}%
\usepackage{listings}%
\usepackage{comment}%
\usepackage{cleveref}%
\usepackage{listings}
\usepackage{comment}
\usepackage{placeins}
\usepackage{xcolor}  
\usepackage[inkscapeformat=pdf]{svg} 
\usepackage{soul} 
\usepackage{wrapfig}
\usepackage{rotating}
\usepackage{longtable}
\usepackage{booktabs}
\usepackage{arydshln}
\usepackage{xurl} 

\newcommand{\q}[1]{``#1''}

\begin{document}

\title[The METRIC-framework]{The METRIC-framework for assessing data quality for trustworthy AI in medicine: a systematic review}

\author*[1]{\fnm{Daniel} \sur{Schwabe}}\email{daniel.schwabe@ptb.de}

\author[1]{\fnm{Katinka} \sur{Becker}}\email{katinka.becker@ptb.de}

\author[1]{\fnm{Martin} \sur{Seyferth}}\email{martin.seyferth@ptb.de}

\author[1]{\fnm{Andreas} \sur{Kla\ss{}}}\email{andreas.klass@ptb.de}

\author[1,2,3]{\fnm{Tobias} \sur{Schaeffter}}\email{tobias.schaeffter@ptb.de}

\affil*[1]{\orgdiv{Division Medical Physics and Metrological Information Technology}, \orgname{Physikalisch-Technische Bundesanstalt}, \orgaddress{\street{Abbestr. 2-12}, \postcode{10587}, \state{Berlin}, \country{Germany}}}

\affil[2]{\orgdiv{Department of Medical Engineering}, \orgname{Technical University Berlin}, \orgaddress{\street{Stra\ss{}e des 17. Juni 135}, \postcode{10623}, \state{Berlin}, \country{Germany}}}

\affil[3]{\orgname{Einstein Centre for Digital Future}, \orgaddress{Wilhelmstra\ss{}e 67}, \postcode{10117}, \state{Berlin}, \country{Germany}}

\abstract{The adoption of machine learning (ML) and, more specifically, deep learning (DL) applications into all major areas of our lives is underway. The development of trustworthy AI is especially important in medicine due to the large implications for patients’ lives. While trustworthiness concerns various aspects including ethical, technical and privacy requirements, we focus on the importance of data quality (training/test) in DL. Since data quality dictates the behaviour of ML products, evaluating data quality will play a key part in the regulatory approval of medical AI products. We perform a systematic review following PRISMA guidelines using the databases PubMed and ACM Digital Library. We identify 2362 studies, out of which 62 records fulfil our eligibility criteria. From this literature, we synthesise the existing knowledge on data quality frameworks and combine it with the perspective of ML applications in medicine. As a result, we propose the METRIC-framework, a specialised data quality framework for medical training data comprising 15 awareness dimensions, along which developers of medical ML applications should investigate a dataset. This knowledge helps to reduce biases as a major source of unfairness, increase robustness, facilitate interpretability and thus lays the foundation for trustworthy AI in medicine. Incorporating such systematic assessment of medical datasets into regulatory approval processes has the potential to accelerate the approval of ML products and builds the basis for new standards.}


\maketitle

\input{sections/section1intro}

\input{sections/section2results}

\input{sections/section3framework}

\input{sections/section4discussion}

\FloatBarrier
\newpage
\input{sections/section5methods}

\FloatBarrier
\newpage
\bibliography{sn-bibliography}

\newpage
\FloatBarrier
\input{sections/section6others}

\end{document}

%% file: sections/section1intro.tex
\section*{Introduction}\label{sec1}

During the last decade, the field of artificial intelligence (AI) and in particular machine learning (ML) has experienced unprecedented advances, largely due to breakthroughs in deep learning (DL)~\cite{vaswani2017attention, deng2018artificial, silver2017mastering, he2016deep, redmon2016you} and increased computational power. Recently, the introduction of easy-to-use yet still extremely capable models such as \mbox{GPT-4}~\cite{OpenAI2023GPT4TR} and Stable Diffusion~\cite{Rombach_2022_CVPR} has further expanded the technology to an even broader audience. The large-scale handling and implementation of AI~\cite{chui2023state} into fields such as manufacturing, agriculture and food, automated driving, smart cities and healthcare has since shifted the topic into the centre of attention of not just scholars and companies but the general public.

The introduction of novel and disruptive technologies is typically accompanied by an oscillating struggle between exploiting technological chances and mitigation of risks. ML is proving to have great potential to improve many aspects of our lives~\cite{esteva2017dermatologist, jumper2021highly, teoh2017rage}. However, the race for implementation and utilisation is currently outpacing comprehension of the technology. The complex and black box character of AI applications has therefore largely steered the public conversation towards safety, security and privacy concerns~\cite{eschenbach2021, aisafetysummit2023}. A lack of confidence of the general population in the transparency of AI prevents its utilisation for society and economic growth. It can lead to a slowed adoption of innovations in crucial areas and discourage innovators from unlocking the technology's full potential. Hence, the demand for regulation (e.g., EU AI Act~\cite{AIAct2023}, US FDA considerations~\cite{food2019proposed}) as well as the need for an improved understanding of AI is ever increasing. This is of particular importance in the field of healthcare due to its large impact on people's lives. The amount of ML solutions in medicine (research tools and commercial products) is steadily on the rise, in particular in the fields of radiology and cardiology~\cite{muehlematter2021approval, zhu20222021}. Despite breakthroughs up to human-level performance~\cite{esteva2017dermatologist, gulshan2016development, ardila2019end, Liu2019}, ML backed medical products are mainly used as diagnosis assistance systems~\cite{zhu20222021} leaving the final decision to medical human professionals. In particular, medical ML solutions are successfully solving the task of image segmentation~\cite{ronneberger2015u, chen2021transunet, Hatamizadeh2022SwinUS}. Due to the unknown consequences of using AI for medical decision-making, more stringent regulatory requirements are of high importance to accelerate the approval process of new AI products into medical practice. Decision-making needs to be supported by reliable health data to generate consistent evidence. One of the drivers for evidence-based medicine approaches was the introduction of scientific standards in clinical practice~\cite{Feinstein1988}. Since then, data integrity (defined by the ALCOA-principles or ALCOA+~\cite{TRS1033}) has become an essential requirement of several guidelines, such as good clinical practice (GCP)~\cite{gcp2016}, good laboratory practice (GLP)~\cite{glp2004} or good manufacturing practice (GMP)~\cite{gmp}. In the pharmaceutical industry, data integrity plays a similarly important role as a requirement for drug trials. While data integrity focuses on maintaining accuracy and consistency of a dataset over its entire life cycle, data quality is concerned with the fitness of data for use.

To improve confidence in AI utilisation in general, the focus is put on the development of so-called trustworthy AI, which aims at overcoming the black box character and developing a better understanding. Several approaches and definitions for trustworthy AI have been discussed and published over the past years by researchers~\cite{adadi2018, Liu2022, Li2023, kale2023, alzubeidi2023}, public entities~\cite{ai2019high, ala2020assessment}, corporations~\cite{DeloitteTrustwortyAI}, and organisations~\cite{vdespecs2022, ignb2022}. Depending on the area of interest, trustworthiness may include (but is far from limited to) topics such as ethics; societal and environmental well-being; security, safety, and privacy; robustness, interpretability and explainability; providing appropriate documentation for transparency and accountability~\cite{adadi2018, Liu2022, Li2023, kale2023, alzubeidi2023, ai2019high, ala2020assessment, DeloitteTrustwortyAI, vdespecs2022, ignb2022}. One of the most critical parts of an AI is the quality of its training data since it has fundamental impact on the resulting system. It lays the foundation and inherently provides limitations for the AI application. If the data used for training a model is bad, the resulting AI will be bad as well (\q{garbage in, garbage out}~\cite{geiger2020garbage}). Neural networks are prone to learning biases from training data and amplifying them at test time~\cite{zhaoamplify2017}, giving rise to a much discussed aspect of AI behaviour: fairness \cite{whittlestone2019ethical}.  Many remedies have been put forward to tackle discriminating and unfair algorithm behaviour~\cite{LearningFairRepresentations, Kim2019, russakovskyfairnessvisual2019}. Yet, one of the main causes of undesirable learned patterns lies in biased training data~\cite{Suresh2021, Mehrabi2021}. Thus, data quality plays a decisive role in the creation of trustworthy AI and assessing the quality of a dataset is of utmost importance to AI developers, as well as regulators and notified bodies.

The scientific investigation of data quality was initiated roughly 30 years ago. The term data quality was famously broken down into so-called data quality dimensions by Wang and Strong in 1996~\cite{Wang1996}. These dimensions represent different characteristics of a dataset which together constitute the quality of the data. Throughout the years, general data quality frameworks have taken advantage of this approach and have produced refined lists of data quality dimensions for various fields of application and types of data. Naturally, this has produced different definitions and understandings. Within this systematic review, we transfer the existing research and knowledge about data quality to the topic of AI in medicine. In particular, we investigate the research question: Along which characteristics should data quality be evaluated when employing a dataset for trustworthy AI in medicine? The systematic comparison of previous studies on data quality combined with the perspective on modern ML enables us to develop a specialised data quality framework for medical training data: the METRIC-framework. It provides a comprehensive list of 15 awareness dimensions which developers of AI medical devices should be mindful of. Knowledge about the composition of medical training data with respect to the dimensions of the METRIC-framework should drastically improve comprehension of the behaviour of ML applications and lead to more trustworthy AI in medicine.

\subsection*{Limitations}
We note that data quality itself is a term used in different settings, with different meanings and varying scopes. For the purpose of this review, we focus on the actual content of a dataset instead of the surrounding technical infrastructure. We do so since the content is the part of a dataset which ML applications use to learn patterns and develop their characteristics. We thus exclude research on data quality considerations and frameworks within the topic of data governance and data management~\cite{khatri2010designing, liaw2014integrated}. This concerns aspects such as data integration~\cite{mo2008method}, information quality management~\cite{lindquist2004data}, ETL processes in data warehouses~\cite{souibgui2019data}, or tools for data warehouses~\cite{gebhardt1998tools, ballou1999enhancing} which do not affect the behavioural characteristics of AI systems. We also omit records discussing case studies of survey data quality~\cite{jenkinson2003cross, lim2008thai}. Furthermore, the METRIC-framework is intended for assessing the quality of a fixed dataset with respect to a specific application. We do not consider assessing the data quality of a dataset in vacuum, nor training strategies to cope with bad data~\cite{candemir2021training, shorten2019survey, feng2021survey, larochelle2009exploring, vincent2008extracting, wang2016non}.

We further point out that the use of the term AI in current discussions is scientifically imprecise since discussions within the healthcare sector almost exclusively revolve around the implementation of ML approaches, in particular of DL approaches. Technically, the term AI spans a much wider range of technologies than just DL as part of the field of ML. Due to the complexity of DL applications and their proficiency in solving tasks deemed to require human intelligence, the terms are currently often used interchangeably in literature. We follow the same vocabulary here (e.g., \q{trustworthy AI}, \q{AI in medicine}) but stress the limitation of our results to ML approaches.\\

%% file: sections/section2results.tex
\section*{Results}\label{sec:history}

\begin{figure}
    \centering
    \includegraphics[width=\textwidth]{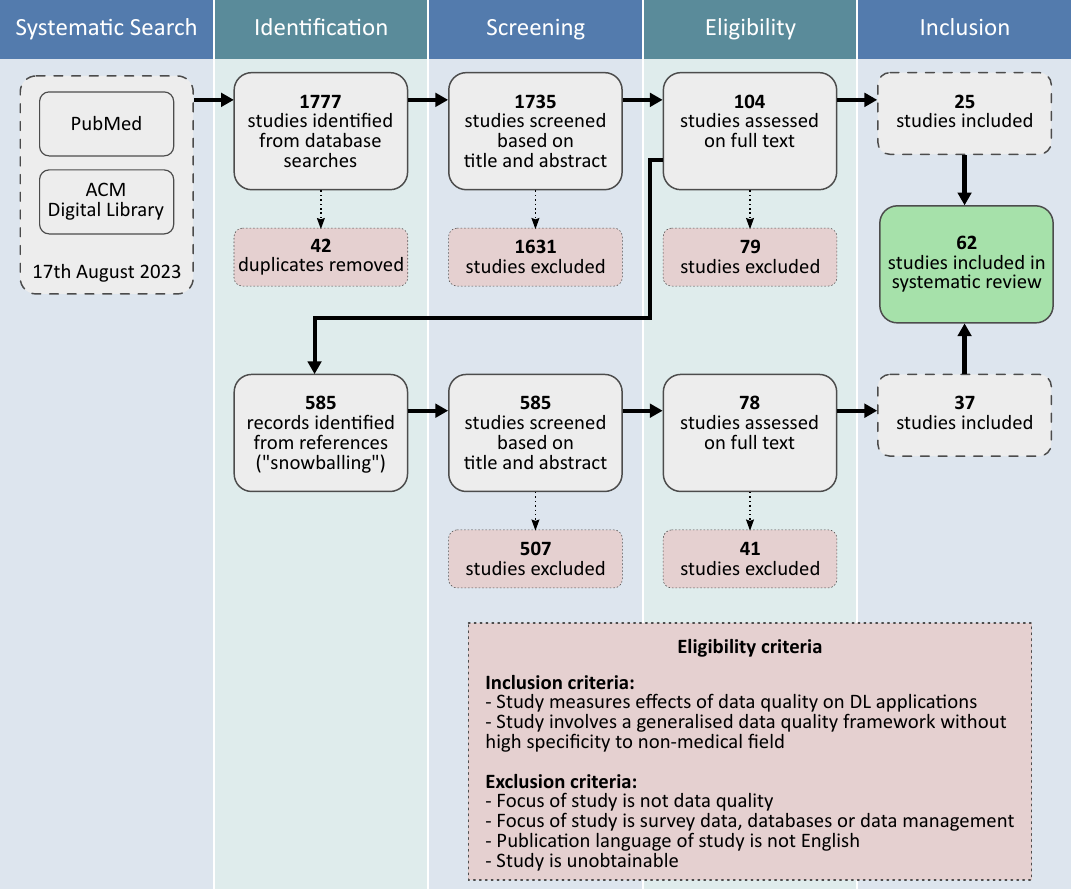}
    \caption{PRISMA flow diagram.
    The flow diagram shows the number of records identified, included and excluded at the different stages of the systematic review. The eligibility criteria for inclusion and exclusion are presented in the bottom right hand side. From a total of 2362 identified studies, the resulting literature corpus on data quality for trustworthy AI in medicine includes 62 studies.}
    \label{fig:prisma}
\end{figure}

In order to answer the research question \q{Along which characteristics should data quality be evaluated when employing a dataset for trustworthy AI in medicine?}, we conducted an unregistered systematic review following the PRISMA guidelines~\cite{page2021prisma}. Our predetermined search string contains variations of the following terms: (i) data quality, (ii) framework or dimensions and (iii) machine learning (see \nameref{sec:methods} for more details and the full search string). The initial search of the databases PubMed and ACM Digital Library was performed on the 17th of August 2023 and yielded 1735 unique results. After title and abstract screening, adding references of the remaining records (\q{snowballing}) and full text assessment, we find 62 records that match our eligibility criteria (see \nameref{sec:methods} and Table~\ref{tab:inex}). This represents the literature corpus that serves as a foundation for answering the research question. The full workflow is illustrated in Figure~\ref{fig:prisma}.

\begin{figure}
    \centering
    \includegraphics[width=\textwidth]{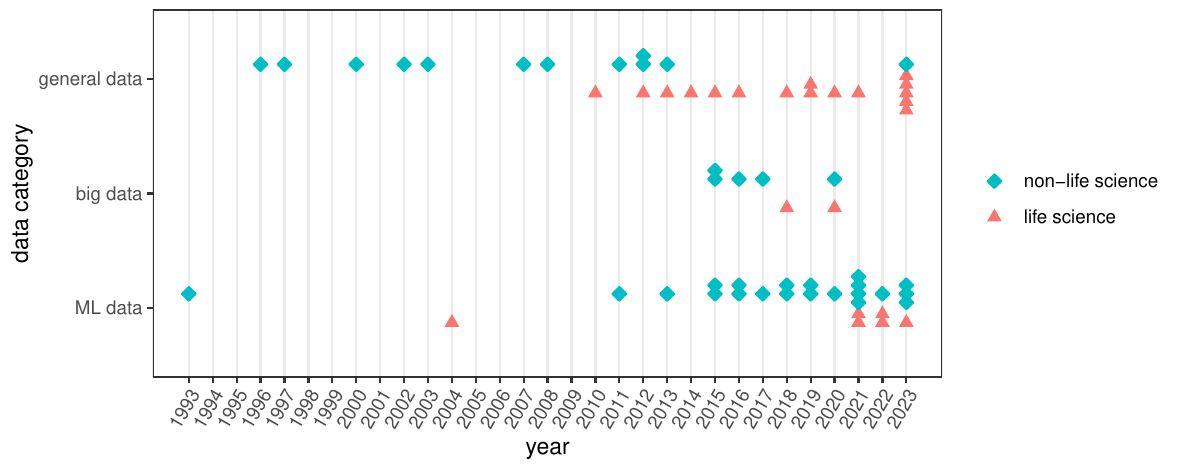}
    \caption{Studies included in our literature corpus sorted by publication date. The 62 studies are divided into the three categories \emph{general data} (28), \emph{big data} (7) and \emph{ML data} (27), which represent major changes in the perception of data quality. The studies' affiliation to non-life science and life science related topics is indicated as well.}
    \label{fig:provoveryears}
\end{figure}

In Figure~\ref{fig:provoveryears}, the papers from our literature corpus are displayed according to their publication year~\cite{Wang1996, Redman1996, Loshin2011, yoon2000managing, Sidi2012, Pipino2002, sebastian2012measuring, Stvilia2007, Kim2003, iso2500025012, dama2013, chan2010electronic, Weiskopf2013, nahm2012data, Chen2014, Bloland2019, Vanbrabant2019, Bian2020, Tahar2023, Kahn2016, Johnson2015, Syed2023, Liu2023, Mashoufi2023, Lewis2023, Schmidt2021, Batini2015, Eder2020, Cai2015, Gao2016, Ramasamy2020, gudivada2017data, Juddoo2018, Qi2021, Cao2020, Bansal1993, twala2013impact, Michel2000, Jouseau2022, che2018recurrent, Benedick2021, Blake2011, He2019, Rolnick2017, Sukhbaatar2014, Wesemeyer2021, Hong2021, sun2017revisiting, LiYang2021, Ranjan2023, Li2021, Fan2022, Pan2023, masko2015impact, Buda2018, Karahan2016, dodge2016understanding, Whang2023, ovadia2019can, Xu2023, shimizu2022effect, Derry2022}. The overarching topics contained in the corpus naturally divide the papers into three categories: \emph{general data} (28 entries), \emph{big data} (7 entries) and \emph{ML data} (27 entries). This reflects the historic development of the research field of data quality during the last 30 years.

\subsection*{General data quality}
The field first shifted into focus with digital and automatically mass-generated data during the 1980s and 1990s causing a need for quality evaluation and control on a broad scale. While during the first 10 years landmark papers~\cite{Wang1996, Redman1996} built the foundation for the field, the most recent 20 years have seen general data quality frameworks published more frequently~\cite{Pipino2002, Loshin2011, yoon2000managing, Stvilia2007, Sidi2012, sebastian2012measuring, Kim2003, dama2013, iso2500025012}. The literature corpus additionally contains general data quality frameworks with high specificity to medical applications~\cite{chan2010electronic, nahm2012data, Weiskopf2013, Chen2014, Johnson2015, Kahn2016, Bloland2019, Vanbrabant2019, Bian2020, Syed2023, Liu2023, Tahar2023, Mashoufi2023, Lewis2023, Schmidt2021} while frameworks with high specificity to non-medical topics~\cite{xu2002data, verma2019data} were excluded.

The early data quality research in the 1980s and 1990s uncovered the lack of objective measures to asses data quality, which led to the introduction of task dependent dimensions and the establishment of a data quality framework from the perspective of the data consumer~\cite{Wang1996}. Another fundamental challenge in the data quality field is the efficient data storage while maintaining quality. This was first investigated with the introduction of a data quality framework from the perspective of the data handler~\cite{Redman1996}. Both approaches to data quality proved to be useful and were unified in one framework~\cite{Loshin2011}. In the following years, the frameworks were further extended~\cite{yoon2000managing, Sidi2012}, equipped with measures~\cite{Pipino2002, sebastian2012measuring} and refined~\cite{Stvilia2007, Kim2003}. Moreover, it became clear that specialised fields such as the medical domain require adapted frameworks.

With the overarching question of how to improve patient care and the rise of electronic health records (EHR) in the 90s of the last century, the need for high data quality in the medical sector increased. The first comprehensive data quality framework for medical data in the literature corpus was established by conducting a survey of quality challenges in EHR~\cite{chan2010electronic}. It considers, among other characteristics, accuracy, completeness and particularly timeliness. However, accuracy is hard to quantify in the medical context as even the diagnosis of experienced practitioners sometimes do not coincide. Accordingly, the notion of concordance of differing data sources was introduced~\cite{Weiskopf2013}. Yet, the data quality frameworks for EHR could only be transferred to other types of medical data to a certain extent. Thus, data quality frameworks for particular data types such as immunisation data, public health data or similar were put forward~\cite{nahm2012data, Chen2014, Bloland2019, Vanbrabant2019, Bian2020, Tahar2023}. The various frameworks still suffered from inconsistent terminology and  attempts were made to harmonise the definitions and assessment~\cite{Kahn2016, Johnson2015, Syed2023, Liu2023, Mashoufi2023, Lewis2023, Schmidt2021}. Particularly, Kahn et al.~\cite{Kahn2016} proposed a framework with exact definitions. While these developments have advanced the understanding of data quality in the context of medical applications, frameworks for EHR frequently focus on the data quality of individual patients~\cite{chan2010electronic, Weiskopf2013}, neglecting data quality aspects for the overall population. In particular, representativeness is often not a factor~\cite{chan2010electronic, Weiskopf2013} while it is a crucial property for secondary use of data in clinical studies \cite{nahm2012data} or when reusing medical data as training data for ML applications.

\subsection*{Big data quality}
As the amount of data from varying sources grew, conventional databases reached their capacity and the field of big data emerged. Big data is generally concerned with handling huge unstructured data streams that need to be processed at a rapid pace, emphasising the need for extended data quality frameworks. This development is reflected by a small wave of papers published between 2015 and 2020~\cite{Batini2015, Cai2015, Juddoo2018, Gao2016, Eder2020, gudivada2017data, Ramasamy2020}. For example, the weaker structure of the data encouraged the use of data quality frameworks that include the data schema as a data quality dimension~\cite{Batini2015, Eder2020}. Further, the increasing amount of data requires the computational efficiency of the surrounding database infrastructure to be a part of big data quality frameworks~\cite{Cai2015, Gao2016, Ramasamy2020}. Computational efficiency is also a limiting factor when ML methods are applied to big data. While it is generally assumed, that more data leads to better results, this has to be balanced with computational capabilities. Hence, a data quality framework was developed that bridges the gap between ML and big data~\cite{gudivada2017data}. We note, that the \q{4 V's} (volume, velocity, veracity and variety) of big data~\cite{laney20013d} implicitly suggest a framework for big data quality. However, the \q{4 V's} are in fact a direct characterisation or definition for when data can be called big data, rather than representing data quality dimensions. They therefore do not contribute to answering our research question and are not further discussed. This might change in the future, when data from wearables or remote patient monitoring sensors become available for health management.

\subsection*{ML data quality}
The performance and behaviour of DL applications heavily depends on the quality of the data used during training as this is the foundation from which patterns are learned. The records of the literature corpus which discuss or empirically evaluate the effect of data quality on DL deal with a wide variety of data types and models. The earliest records starting in the 1990s employ simple neural network (NN) designs and investigate data quality of tabular data~\cite{Michel2000, Bansal1993, twala2013impact, Qi2021, Cao2020, Jouseau2022}, while records from the 2010s and 2020s increasingly use advanced DL architectures and look at data quality in the context of time series~\cite{che2018recurrent, Benedick2021, Blake2011}, images~\cite{Sukhbaatar2014, Rolnick2017, Karahan2016, dodge2016understanding, Buda2018, LiYang2021, Hong2021, sun2017revisiting, Fan2022, Ranjan2023, Whang2023, masko2015impact, He2019, Pan2023, ovadia2019can, Li2021, Wesemeyer2021, Jouseau2022}, natural language~\cite{ovadia2019can, Whang2023, shimizu2022effect, Xu2023} or molecular data~\cite{Derry2022, Fan2022}.

Contrary to the big data and general data quality literature from our corpus, the DL papers focus on the evaluation of one or very few specific data quality dimensions without (yet) considering broader theoretical data quality frameworks. Dimensions that are predominantly investigated are those which can easily be manipulated and lend themselves to be applicable to a wide range of datasets irrespective of specific tasks. The most prominent dimension is \textit{amount of data}~\cite{shimizu2022effect, Li2021, LiYang2021, Derry2022, Hong2021, Michel2000, sun2017revisiting, He2019, Fan2022, Ranjan2023, Sukhbaatar2014, Wesemeyer2021, Xu2023} which is empirically shown to benefit performance, albeit in a saturating manner. Another dominant topic is completeness to which the ML community almost exclusively refers to as \textit{missing data}~\cite{Blake2011, Benedick2021, Jouseau2022, Michel2000, Whang2023, che2018recurrent}. The effect that data errors have on the DL application is also frequently investigated. Specifically, this is done by separately looking at \textit{contaminated features} (inputs of a NN)~\cite{Blake2011, Bansal1993, Whang2023, He2019, Benedick2021, twala2013impact, Karahan2016, dodge2016understanding} and \textit{noisy targets} (predictions generated by a NN)~\cite{Blake2011, Whang2023, He2019, Sukhbaatar2014, Rolnick2017, twala2013impact, Wesemeyer2021, Xu2023}. Many ML settings are classification tasks which is reflected by the corpus often addressing label noise~\cite{Whang2023, He2019, Sukhbaatar2014, Rolnick2017, Wesemeyer2021, Xu2023}. One record highlights the hefty weight that physicians’ annotations carry in medicine~\cite{Wesemeyer2021}. In order to evaluate the effect of data quality (features or targets) on ML applications, the training data is commonly manipulated. On the feature (input) side, e.g., images are distorted by adjusting contrast whereas time series sequences are disturbed by swapping elements. On the target side, e.g., correct labels are randomly replaced by false ones.

When it comes to the concrete behaviour change of the DL algorithm, most of the DL papers in the literature corpus investigate the robustness of a model, i.e. the stable behaviour of a model when facing erroneous or a limited amount of inputs. Only few records investigate generalisability~\cite{ovadia2019can, Pan2023, Jouseau2022}, a model's capability of coping with new, unseen data. Another noteworthy exception is Ovadia et al.~\cite{ovadia2019can} who additionally study predictive uncertainty.

Overall, theoretical data quality frameworks enjoy little attention by the ML community due to the novelty of the ML research field. Papers often focus on few specific data quality dimensions and tasks. Each task comes with its specific data type, necessitating different approaches to manipulate the data and measure these effects. The research dealing with the impact of manipulated data is heavily skewed towards robust behaviour in the sense of predictive performance. Other possibly affected aspects such as explainability or fairness are underrepresented and to some degree neglected which is a potential shortcoming for safety-critical applications such as medical diagnosis predictions.

%% file: sections/section3framework.tex
\section*{METRIC-framework for medical training data}\label{sec:results}

\begin{figure}
    \centering
    \includegraphics[width=0.98\textwidth]{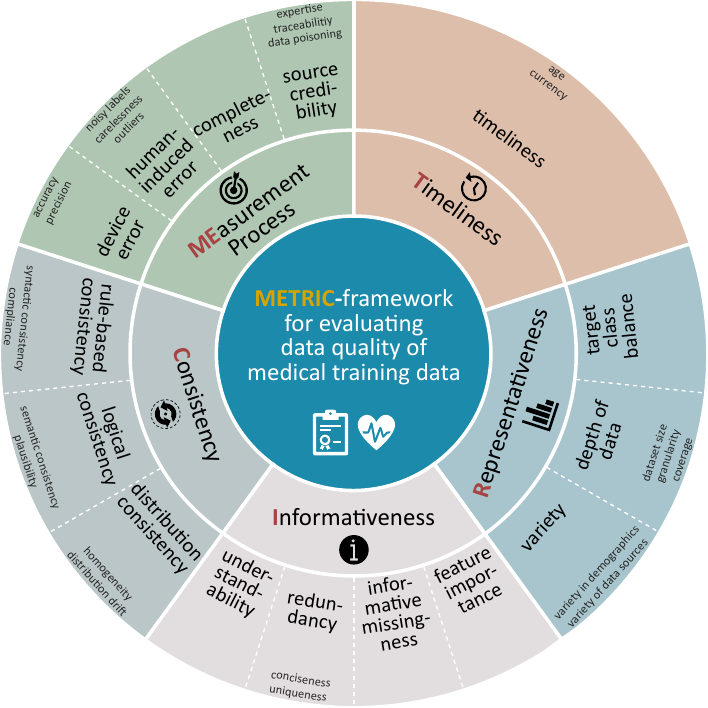}
    \caption{The METRIC-framework. This specialised framework for evaluating data quality of medical training data includes a comprehensive set of awareness dimensions. 
    The inner circle divides data quality into five clusters. These clusters contain a total of 15 data quality dimensions, which are shown on the outer circle. The subdimensions presented in gray on the border of the figure contribute to the superordinate dimension. Due to the shape of the graphic, we refer to it as \emph{wheel of data quality}.}
    \label{fig:dqframework}
\end{figure}

The literature corpus has shown that while similar ideas exist for the assessment of data quality across fields and applications, the idiosyncrasy of each field or application can only be captured by specialised frameworks rather than by a one-model-fits-all framework. The evaluation of data quality plays a particularly important role in the field of ML due to the fact that its behaviour is not only dependent on the algorithm choice but also strongly depends on its training data. At the same time, ML is implemented in various fields, each processing and requiring different types and qualities of data. We therefore propose a specialised data quality framework for evaluating the quality of medical training data: the METRIC-framework (Figure~\ref{fig:dqframework}), which is based on our literature corpus~\cite{Wang1996, Redman1996, Pipino2002, Loshin2011, yoon2000managing, Stvilia2007, Sidi2012, sebastian2012measuring, Kim2003, dama2013, iso2500025012, chan2010electronic, Weiskopf2013, nahm2012data, Chen2014, Bloland2019, Vanbrabant2019, Bian2020, Tahar2023, Kahn2016, Johnson2015, Syed2023, Liu2023, Mashoufi2023, Batini2015, Cai2015, Juddoo2018, Gao2016, Eder2020, gudivada2017data, Ramasamy2020, Michel2000, Bansal1993, twala2013impact, Qi2021, Cao2020, che2018recurrent, Benedick2021, Sukhbaatar2014, Rolnick2017, Karahan2016, dodge2016understanding, LiYang2021, Hong2021, sun2017revisiting, Fan2022, Ranjan2023, Whang2023, masko2015impact, Buda2018, He2019, Pan2023, ovadia2019can, Li2021, shimizu2022effect, Derry2022, Jouseau2022, Lewis2023, Schmidt2021, Blake2011, Wesemeyer2021, Xu2023}. From now on, we refer to data quality for training data of medical ML applications only. We point out that our framework does not yet include a guideline on the assessment or measurement of data qualities but rather presents a set of awareness dimensions which play a central role in the evaluation of data quality. Additionally, we note that the METRIC-framework is specifically not designed to assess the data quality of a dataset in vacuum, nor with respect to all potential ML applications the dataset might be used for. Rather, the intention of the METRIC-framework is to assess the appropriateness of a dataset with respect to a specific and known purpose of the desired medical AI.

While examining the literature corpus, we found that terms describing data quality appear under varying definitions, or often with no definition at all. While standardisation efforts exist for the terminology in the context of evaluating data quality~\cite{dama2020, iso2500025012, ieee1990ieee}, they are often not employed or did not exist yet for older papers making comparisons difficult. Therefore as a first step, we extracted all mentioned data quality dimensions from the literature corpus together with their definitions (if present) and added them to a list. This yielded 383 different terms with 637 mentions across all papers. Second, we hierarchically clustered the terms with respect to their intended meaning and according to their dependencies into clusters, dimensions and subdimensions (see \nameref{sec:methods} for more details on data extraction). We thus obtained 38 relevant dimensions and subdimensions which are displayed on the outer circle of Figure~\ref{fig:dqframework}. In Tables~\ref{tab:dqd_to_papers1}--\ref{tab:dqd_to_papers6}, we provide a complete list of definitions for all 38 relevant dimensions and subdimensions, as well as their hierarchy and references with respect to the literature corpus. We adopted definitions from a recent data quality glossary~\cite{dama2020} if they existed there and met our understanding of the dimension in the given context of medical training data. If necessary, we included definitions given by Wang et al.~\cite{Wang1996} in a second iteration. If none of these two sources suggested an appropriate definition, we captured the meaning of the desired term on the basis of the literature corpus and thus determined its definition in the context of medical training data.

\begin{wrapfigure}{r}{0.45\textwidth}
    \centering
    \includegraphics[width=0.44\textwidth]{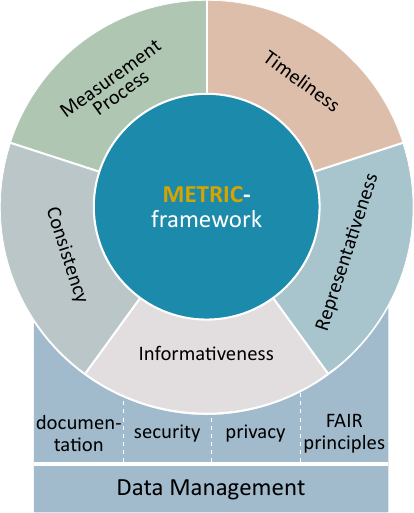}
    \caption{The cluster \emph{data management} is concerned with the effective usage of the dataset. It includes basic requirements for the dataset but does not address data quality issues regarding its content. Therefore, they can be seen as a prerequisite for assessment using the METRIC-framework. Figuratively speaking, the \emph{data management} cluster serves as a stable foundation for the wheel of data quality.}
    \label{fig:suitability}
\end{wrapfigure}

The METRIC-framework encompasses three levels of details: \emph{clusters} which pool similar dimensions; \emph{dimensions} which are individual characteristics of data quality; and \emph{subdimensions} which split larger dimensions into more detailed attributes (compare Figure~\ref{fig:dqframework} from inside to outside). Besides the terms contained in the METRIC-framework, we found several frequently mentioned dataset properties which we, for our purpose, want to separate from the METRIC-framework. We summarise these additional properties under a separate cluster called \emph{data management} (Figure~\ref{fig:suitability}). The attributes included in this cluster ensure that a dataset is well-documented and effectively usable. In particular, it includes the properties \emph{documentation}, \emph{security} and \emph{privacy}, as well as the well-established \emph{FAIR-Principles}~\cite{Wilkinson2016} requiring data to be findable, accessible, interoperable and reusable. While these properties play a central role in the handling of data, the METRIC-framework is targeted at the content of a dataset. Therefore, we see the \emph{data management} cluster as a prerequisite for data quality assessment by the METRIC-framework which itself divides the concept of data quality for the content of a dataset into five clusters: \emph{measurement process}, \emph{timeliness}, \emph{representativeness}, \emph{informativeness}, \emph{consistency}. A summary of the characteristics and key aspects of all five clusters is given in Table~\ref{tab:clusters}.

\begin{table}[h]
\begin{tabular}{ l p{8cm} }
  Cluster & Description\\
  \hline
  \textbf{ME}easurement process & 
  Concerned with technical or human influences that affect the data acquisition process. Includes \emph{device errors} leading to poor accuracy and precision,  \emph{human-induced errors} such as noisy labels, \emph{completeness} counting the number of missing values, and \emph{source credibility} estimating the reliability of the data. \\[5pt]
  \textbf{T}imeliness &  
  Concerned with changes in time and whether they are appropriately reflected in the dataset. Includes \emph{age} concerned with the relation between creation date and the date of usage, as well as \emph{currency} representing the updatedness of the data.\\[5pt]
  \textbf{R}epresentativeness & 
  Concerned with the appropriate and comprehensive representation of the targeted population in the dataset. Includes \emph{variety} indicating whether a sufficiently broad range of demographics and data sources is present, \emph{depth of data} investigating whether the overall amount of data as well as coverage of subpopulations is sufficient, and \emph{target class balance} judging whether classes of the target variable are appropriately sized for use in ML.\\[5pt]
  \textbf{I}nformativeness &
  Concerned with how well the data conveys the information it describes. Includes \emph{understandability} indicating whether data is easily comprehended and without ambiguity, \emph{redundancy} investigating whether information is provided more than once, \emph{informative missingness} quantifying whether missing values carry additional information, and \emph{feature importance} estimating the value features add for the application.\\[5pt]
  \textbf{C}onsistency & 
  Concerned with the consistency of presentation and composition of the dataset. Includes \emph{rule-based consistency} investigating whether data is presented in a consistent format that follows internal and external rules, \emph{logical consistency} judging whether data is logically sound and without contradictions, and \emph{distribution consistency} considering whether different subpopulations have similar statistical properties.\\
\end{tabular}
\caption{The key characteristics of each of the five clusters of the METRIC-framework. For more detail see the main text and the definitions of dimensions and subdimensions in Tables~\ref{tab:dqd_to_papers1}--\ref{tab:dqd_to_papers6}.}\label{tab:clusters}
\end{table}

\subsection*{Measurement process}
The cluster \emph{measurement process} captures factors that influence uncertainty during the data acquisition process. Two of the dimensions within this cluster differentiate between technical errors originating from devices during measurement (see \emph{device error}) and errors induced by humans during, e.g.,data handling, feature selection or data labelling (see \emph{human-induced error}). For the dimension \emph{device error}, we distinguish between the subdimension \emph{accuracy}, the systematic deviation from the ground truth (also called bias), and the subdimension \emph{precision}, the variance of the data around a mean value (also called noise). In practice, a ground truth for medical data is most often not attainable, making \emph{accuracy} evaluation impossible. In that case, the level and structure of noise in the training data should be compared to the expected noise in the data after AI deployment. If the training data only contains low noise but the AI is utilised in clinical practice on data with much higher noise levels, the performance of the AI application might not be sufficient since the model did not face suitable error characteristics during training. Therefore, lower noise data is not necessarily better and adding noise to the training data might in some instances even improve performance~\cite{Bishop1995, Grandvalet1997, smilkov2017smoothgrad}. The errors belonging to the dimension \emph{human-induced errors} are of a fundamentally different nature and need to be treated accordingly. This type of error includes human \emph{carelessness} and \emph{outliers} in the dataset due to (unintentional) human mistakes. The final subdimension, \emph{noisy labels}, is one of the most relevant topics in current ML research~\cite{Whang2023, He2019, Sukhbaatar2014, Rolnick2017}. Since in the medical domain, supervised learning paradigms are prevalent, proper feature selection and reliable labelling are indispensable. However, human decision making can be highly irrational and subjective, especially in the medical context~\cite{thaler2009nudge, kahneman2011thinking, malossini2006detecting}, representing one of various sources of labelling noise~\cite{frenay2013classification}. Among expert annotators there is often considerable variability~\cite{frenay2013classification, menze2014multimodal}. Even in the most common (non-medical) datasets of ML (e.g., MNIST~\cite{deng2012mnist}, \mbox{CIFAR-100}~\cite{krizhevsky2009learning}, \mbox{Fashion-MNIST}~\cite{xiao2017fashion}) there is a significant percentage of wrong labels~\cite{muller2019identifying, northcutt2021confident}. In contrast to the precision of instruments, noise in human judgements is demanding to be assessed through so-called noise audits to identify different factors, like pattern noise and occasion noise in the medical decision process~\cite{Kahneman2021}. Such intra- and inter-observer variability has always been a highly important topic in many medical disciplines, e.g., in radiology where guidelines, training and consensus reading approaches are used to reduce noise~\cite{Jarmillo2022}. 

Another issue that frequently occurs in the data acquisition process and which plays an important role in ML is the absence of data values with unknown reason. We follow the ML vocabulary by capturing this quality issue with the dimension \emph{completeness}. It is usually measured by the ratio of missing to total values. Apart from the mostly quantitative dimensions within the cluster, the dimension \emph{source credibility} is concerned with mostly qualitative characteristics. On the one hand, it includes the question whether or not the measured data can be trusted based on the \emph{expertise} of people involved in data measurement, processing and handling. On the other hand, the subdimension \emph{traceability} evaluates whether changes from original data to its current state are documented. Being aware of modifications such as the exclusion of outliers or data normalisation and their chosen algorithms are necessary for understanding the composition of the data. Finally, the subdimension \emph{data poisoning} considers whether the data was intentionally corrupted (e.g., adversarial attacks) to cause distorted outcomes. The entire cluster \emph{measurement process} is crucial for data quality evaluation in the medical field since errors may propagate through the ML model and lead to false diagnosis or treatment of patients.

\subsection*{Timeliness}
Since medical knowledge and understanding is subject to constant development, it is important to investigate the cluster \emph{timeliness} which indicates whether the point in time at which the dataset is used in relation to the point in time at which it was created and updated is appropriate for the task at hand. Indications for diagnoses based on medical data may have changed since a dataset was created and labelled, and changes in coding systems (such as the transition from ICD-9 to ICD-10 or ICD-9-CM to ICD-10-CM) may affect mortality and injury statistics~\cite{anderson2001comparability, Sebastio2021}). The \emph{age} of the data dictates whether such investigations are necessary. In such cases, the labels or standards utilised would then have to be appropriately updated to satisfy the subdimension \emph{currency}. Furthermore, knowledge about the subdimension \emph{age} might provide information about precision and accuracy of the measurement as it gives insight into the technology used during data acquisition.

\subsection*{Representativeness}
Another central cluster, especially for medical applications, is \emph{representativeness}. Its dimensions are concerned with the extent to which the dataset represents the targeted population (such as patients) for which the application is intended. Whether the population of the dataset covers a sufficient range in terms of age, sex, race or other background information is the topic of the subdimension \emph{variety in demographics} contained within the dimension \emph{variety}. This dimension also contains the subdimension \emph{variety of data sources} concerned with questions such as: Does the data originate from a single site? Were the measurements done with devices from the same or different manufacturers? Appropriately investigating such questions can provide a strong indication for the applicability and generalisability of the ML application in different environments~\cite{remedios2020distributed, onofrey2019generalizable,pooch2020can, glockermultisite2019}. The dimension \emph{depth of data} is one of the main topics of the ML papers in our literature corpus. Apart from the subdimension \emph{dataset size} already discussed in the previous section, this dimension also includes the subdimension \emph{granularity}, which considers whether the level of detail (e.g., the resolution of image data) is sufficient for the application, as well as the subdimension \emph{coverage}, which investigates whether sub-populations (e.g., specific age groups) are still diverse by themselves (e.g., still contain all possible diagnoses in case of classification applications). Finally, the highly-discussed dimension \emph{target class balance} pays tribute to the technical requirements of ML \cite{Buda2018,LiYang2021,masko2015impact,Pan2023,He2019}. An algorithm must learn patterns for specific classes from the training data. However, strong imbalances in the class ratio could be caused by, e.g., rare diseases. In order to still be able to properly learn corresponding patterns it may be helpful to deliberately overrepresent rare classes in the dataset instead of matching their real world distribution~\cite{Chawla2002, HaiboHe2008}.

\subsection*{Informativeness}
The cluster \emph{informativeness} considers the connection between the data and the information it provides and whether the data does so in a clear, compact and beneficial way. First of all, the \emph{understandability} of the data considers whether the information of the data is easily comprehended. Second, the dimension \emph{redundancy} investigates whether such information is concisely communicated (see subdimension \emph{conciseness}) or whether redundant information is present such as duplicate records (see subdimension \emph{uniqueness}). The dimension \emph{informative missingness} answers the question whether the patterns of missing values provide additional information. Che et al.~\cite{che2018recurrent} find an informative pattern in the case of the \mbox{MIMIC-III} critical care dataset~\cite{johnson2016mimic} which displays a correlation between missing rates of variables and ICD9-diagnosis labels. Missingness patterns are categorised by the literature into either \textit{not missing at random} (NMAR), \textit{missing at random} (MAR) or \textit{missing completely at random} (MCAR)~\cite{rubin1976inference, schafer2002missing}. Finally, \emph{feature importance} is concerned with the overall relevance of the features for the task at hand and moreover with the value each feature provides for the performance of a ML application since the quantity of data has to be balanced with computational capability. Valuable features might in many cases be as important as dataset size~\cite{mazumder2022dataperf}, which is a frequently discussed topic in the data-centric AI community~\cite{zhadatacentricsurvey2023}.

\subsection*{Consistency}
The dimensions belonging to the cluster \emph{consistency} illuminate the topic of consistent data presentation from three perspectives. \emph{Rule-based consistency} summarises subdimensions concerned with format (\emph{syntactic consistency}), which includes the fundamental and well-discussed topic of data schema~\cite{Batini2015}, and the conformity to standards and laws (\emph{compliance}). These subdimensions ensure that the dataset is easily processable on the one hand and comparable and legally correct on the other. \emph{Logical consistency} evaluates whether or not the content of the dataset is free of contradictions, both within the dataset (e.g., a patient without kidneys that is diagnosed with kidney stones) and in relationship to real world knowledge (e.g., a 200-year-old patient). The last dimension of the cluster, \emph{distribution consistency}, concerns the distributions and their statistical properties of relevant subsets of the total dataset. While the subdimension \emph{homogeneity} evaluates whether subsets have similar or different statistical properties at the same point in time (e.g., can data from different hospitals be identified by statistics?), the subdimension \emph{distribution drift} deals with varying distributions at different time points. This subdimension can be neglected if the dataset is not continuously changing over time, but distribution drift is sometimes unconsciously discarded due to a lack of model surveillance. Therefore, it is a prominent research topic~\cite{ovadia2019can} and the unconsciousness furthermore underlines the importance of \emph{distribution drift} for medical applications~\cite{Bian2020}.

%% file: sections/section4discussion.tex
\section*{Discussion}\label{sec:discussion}

The METRIC-framework (Figure~\ref{fig:dqframework}) represents a comprehensive system of data quality dimensions for medical training data with respect to a specific task. We stress again that these dimensions should be regarded as awareness dimensions. They provide a guideline along which developers should familiarise themselves with their data. Knowledge about these characteristics is helpful for recognising the reason for the behaviour of an AI system. Understanding this connection enables developers to improve data acquisition and selection which may help in reducing biases, increasing robustness, facilitating interpretability and thus has the potential to drastically improve the AI's trustworthiness.

With training data being the basis for almost all medical AI applications, the assessment of its quality gains more and more attention. However, we note that providing a division of the term data quality into data quality dimensions is only the first step on the way to overall data quality assessment. The next step will be to equip each data quality dimension with quantitative or qualitative measures to describe their state. The result of this measure then has to be evaluated with respect to the question: Is the state of the dimension appropriate for the desired AI algorithm and its application? These three steps (choosing a measure, obtaining a result, evaluating its appropriateness for a task) can be applied to each dimension and subdimension. Appropriately combining the individual outcomes can potentially serve as a basis for a measure of the overall data quality in future work.

So far the dimensions in the METRIC-framework are not ranked in any way. However, it is clear that some of them are more important than others. Therefore, some dimensions deserve more attention in the assessment process or might even be a criterion for exclusion of a dataset for a certain task. These dimensions should be among the first to be assessed in practice. On the other hand, some dimensions are much more difficult to measure and evaluate than others. This can be due to their qualitative nature, the complexity of the statistical measure, the degree of use-case dependence or the expert knowledge that is needed for the assessment, to name a few. These considerations are of central interest for the development of a complete data quality assessment and examination process.

\begin{figure}
    \centering
    \includegraphics[width=0.95\textwidth]{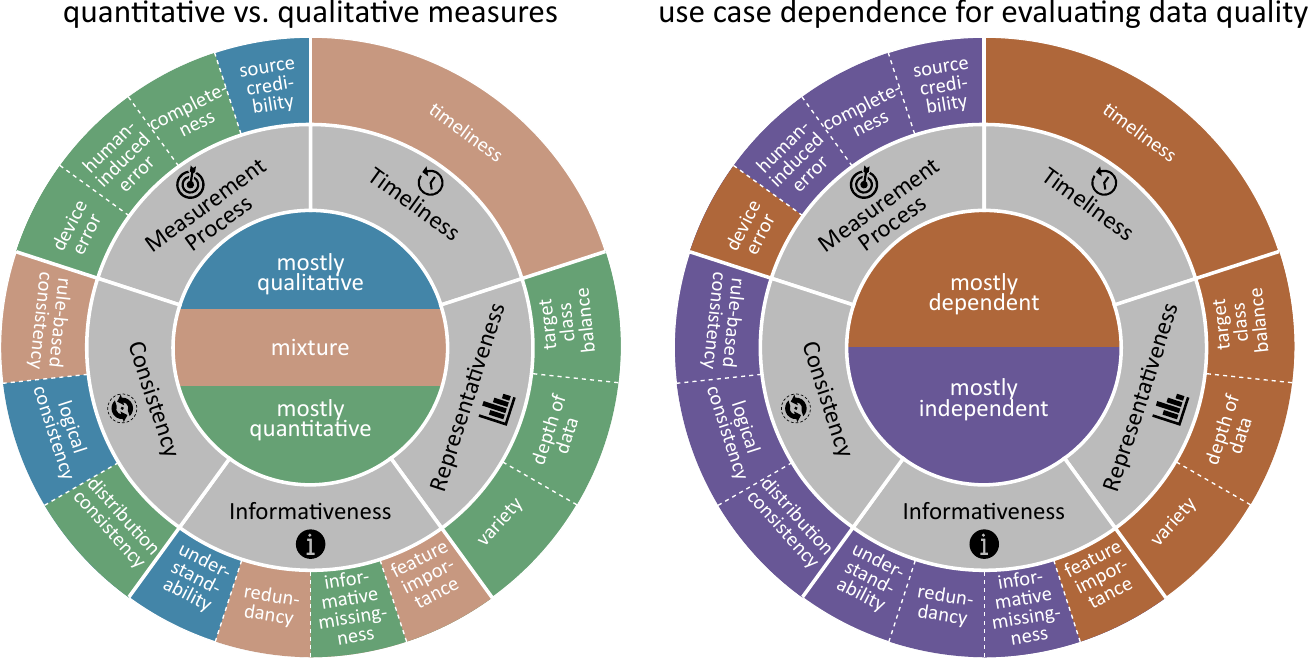}
    \caption{Categorisation of dimensions along the properties \emph{quantitative vs. qualitative measure} (left) and \emph{use case dependence for evaluating data quality} (right). The affiliation to a category is colour-coded. The colour scale is presented in the inner circle.}
    \label{fig:measures}
\end{figure}

In Figure~\ref{fig:measures}, we provide insights that should be taken into consideration when practically assessing data quality. We classify each of the 15 awareness dimensions along two different properties. On the one hand, we estimate whether a dimension requires mostly \emph{quantitative} or \emph{qualitative} measures. We observe that about half of the dimensions require mostly quantitative measures while a fifth necessitate more manual inspection by qualitative measures (see left-hand side of Figure~\ref{fig:measures}). Being able to choose quantitative measures typically implies more objectivity and enables automation, two desirable properties for quality assessment. Dimensions categorised as mostly qualitatively measurable or requiring a mixture of quantitative and qualitative input will typically require specific domain knowledge from the medical field. Such domain knowledge can be difficult to obtain and expensive.

On the other hand, we consider whether the state of a dimension or the evaluation of its appropriateness level is \emph{use case dependent} (see right-hand side of Figure~\ref{fig:measures}). This is of interest to developers as use case dependent dimensions require not only additional knowledge, work and time during quality assessment but also during quality improvement of data. Our findings suggest a clear division of the wheel of data quality after categorising all 15 dimensions. The clusters \emph{representativeness} and \emph{timeliness} as well as the dimensions \emph{device error} and \emph{feature importance} belong to the group of use case dependent dimensions. Whether a dataset is representative of the targeted population can only be evaluated with knowledge of the use case. Similarly, the importance of features changes between applications. Whether the age and currency of the data (see dimension \emph{timeliness}) are appropriate can also differ depending on the task. For instance, the coding standard the data should conform to depends on the application. The newest standards are not necessarily the best if in practice these standards are not implemented (see section on \emph{Timeliness}). Similarly, reducing noise levels in the data is not necessarily better for all applications. It rather depends on the expected noise levels of the application (see section on \emph{Measurement process} for more detail).

For an overall assessment of the quality of the dataset, we estimate that on average the dimensions of the \emph{representativeness} cluster together with the dimensions \emph{feature importance}, \emph{distribution consistency} and \emph{human-induced error} are crucial factors. Ignoring a single one of these dimensions potentially has proportionally larger effects on the AI application than other dimensions. This might also depend on type of ML problem. Actual quantification of the effect of data quality dimensions on ML applications is part of ongoing and future research. Nevertheless, we for now recommend prioritising these six dimensions if it is possible to dedicate time to evaluating or improving a dataset. With the exception of the dimension \emph{feature importance}, all of the crucial dimensions are simultaneously measured mostly quantitatively making them primary candidates for software tools designed for improving the quality of datasets.

The importance of data quality for medical AI products is undisputed and gaining more and more attention with on-going discussions about fairness and trustworthiness. Parts of future regulation and certification guidelines will not only include AI algorithms but likely also require evaluating the quality of datasets used for their training and testing. Such inclusion of data quality in regulation requires systematic assessment of medical datasets for which the METRIC-framework and our additional considerations can serve as a starting point. This has the potential to accelerate the approval of new ML products into medical practice.

%% file: sections/section5methods.tex
\section*{Methods}\label{sec:methods}
(Supplementary Information)
\begingroup
\renewcommand{\thetable}{S\arabic{table}}

\subsection*{Literature review}
In order to answer the research question \q{Along which characteristics should data quality be evaluated when employing a dataset for trustworthy AI in medicine?}, we conducted a systematic review following the PRISMA guidelines~\cite{page2021prisma}. The goal of such a review is to objectively collect the knowledge of a chosen research area by summarising, condensing and expanding the ideas to further its progress. PRISMA reviews commonly follow four main steps: (i) Searching suitable databases with carefully formulated search strings and extracting matching papers; (ii) screening titles and abstracts to include or exclude papers based on predetermined criteria; (iii) extending the literature list by screening titles and abstracts of all referenced papers from the included papers (called \q{snowballing}); (iv) screening the full text of all still included papers with respect to the eligibility criteria to build the final literature corpus.

\subsection*{Search strategy}
Our research question aims at combining the knowledge from the field of general data quality frameworks with insights about the effects that the quality of training data has on ML applications in medicine. This should ultimately lead to a novel framework for data quality in the context of medical training data. Therefore, we built a search string that on the one hand targeted papers about data quality frameworks by combining variations of \q{data quality} with variations of the terms \q{framework} and \q{dimensions}. On the other hand, we attempted to collect papers about the connection between the quality of training data and the behaviour of a DL application by again combining variations of the word \q{data quality} but this time with variations of \q{machine learning}, including \q{artificial intelligence} and \q{deep learning} (for the exact search query, see \nameref{sstring}). We then performed the database search on two thematically suitable online databases: PubMed and ACM Digital Library. All retrieved results were concatenated and duplicates removed, yielding 1735 records.

\subsection*{Search query}\label{sstring}
The following search string in pseudo code was executed on the 17th of August 2023 on PubMed and ACM Digital Library:
\begin{verbatim}
(("data quality" OR "data-quality" 
  OR "data qualities" OR "quality of data" 
  OR  "quality of the data" OR "qualities of data"
  OR "qualities of the data"
 )
 AND 
 ("dimension" OR "dimensions"
  OR "AI" OR "artificial intelligence"
  OR "ML" OR "machine learning" 
  OR "deep learning"
  OR "neural network" OR "neural networks"
 )
)
OR
("data quality framework" OR "data quality frameworks"
 OR "framework of data quality" OR "framework for data quality"
)
\end{verbatim}

The chosen databases supported exact (instead of fuzzy) searches, expressed by quotation marks around keywords. The search was applied to the title and abstract fields of all records of the databases.

\subsection*{Eligibility criteria}
In Table~\ref{tab:inex}, our chosen eligibility criteria that were applied to the various screening steps are listed. Papers were included if they either provided broad-scale data quality frameworks with general purpose or with specificity to a medical application, or if they discussed or quantified the effects of at least one training data quality dimension on DL behaviour. In contrast, papers were excluded if they (i) either discussed frameworks with specificity to non-medical fields or (ii) only considered single or few data quality dimensions without reference to ML or (iii) focused on the quality of data management and surveys. No limits were imposed with respect to publication date or publisher source (i.e. peer-reviewed or not), while non-English records and inaccessible records were omitted.

\input{tables/inclusion_exclusion}

\subsection*{Literature review process}
Titles and abstracts from the records of the database search were screened with respect to the eligibility criteria. This was done by two authors independently to mitigate biases. In case of disagreement, consensus was achieved by discussion. If necessary, a third author was consulted to arrive at the final decision. This step reduced the number of records to 104.

The snowballing step expands the scope of the literature corpus to make it more independent of the initially chosen databases and search string which is important to reduce bias. For the process of snowballing, we considered all references from the so far 104 included papers which resulted in adding 585 records to the literature list. Analogously, title and abstract screening was performed on these new entries with the same criteria and workflow as before, leaving 78 additional papers from snowballing.

As a final step, all 182 remaining papers were screened on the full text with respect to the eligibility criteria. In the end, 62 entries passed all screening steps.

For each retrieved record, the decision whether to include or exclude was documented along with the corresponding eligibility criterion.

\subsection*{Data extraction strategy}\label{sec:dataextraction}
In order to introduce a comprehensive data quality framework, the 62 selected records were each read by two authors and all terms that were deemed relevant to describe data quality were extracted. We discarded terms if (i) their scope is limited to a specialised data source and not transferable to a general framework, (ii) the term refers to the quality of database infrastructure or (iii) no definition was given and it was impossible to grasp the intended meaning from the context. The accepted terms were copied into an Excel sheet, which served as a starting template for the METRIC-framework. We clustered related concepts into groups according to the terms' definition or intended meaning. From these small and detailed groups we formed the so-called subdimensions, ensuring that each subdimension is mentioned by at least three references in the literature corpus, otherwise the level of detail was deemed too great leading to further grouping. Due to the emphasis on ML applications, we initially also included subdimensions that cover concepts which are well-accepted to be influential in the ML community but have not yet been the topic of systematic quantification and therefore have fewer references. After all further steps, this now only applies to the subdimension \emph{variety of data sources}.

Thorough discussion of all authors about underlying concepts and definitions of the subdimensions resulted in hierarchically grouping these into dimensions and the dimensions into clusters. In parallel to this grouping, all authors reached consensus on definitions for dimensions and subdimensions of the METRIC-framework. The definitions were adopted from a recent data quality glossary~\cite{dama2020} if they existed there and met our understanding of the vocabulary in the given context of medical training data. If necessary, we included definitions given by Wang et al.~\cite{Wang1996} in a second iteration. If none of these two sources suggested an appropriate definition, we captured the meaning of the desired term on the basis of the literature corpus and thus determined its definition in the context of medical training data.

\FloatBarrier
\newpage
\input{tables/RecordsByDataType}

\FloatBarrier
\newpage
\subsection*{Definitions of data quality dimensions}\label{sec:dqdtopapers}
\input{tables/DQD_to_papers}

%% file: tables/inclusion_exclusion.tex
\begin{table}[t]
\begin{tabular}{p{0.49\textwidth} p{0.49\textwidth}}
Inclusion criteria                                           &           \\ \hline
Criterion                                                    & Reasoning \\ \hline
 Study measures effects of data quality on DL applications & Get an insight into what dimensions are being investigated by the literature         \\
Study involves a generalised data quality framework without
high specificity to non-medical field                          &  Incorporate earlier work on data quality aspects        \\ \hline
                                                             &           \\
Exclusion criteria                                           &           \\ \hline
Criterion                                                    & Reasoning \\ \hline
Focus of study is not data quality                           &  Not related to research question         \\
Focus of study is survey data, databases or data management                      & Out of scope w.r.t. our research question; too specific topics without accounting for task-specific content          \\
Publication language of study is not English                             & /    \\
Study is unobtainable                            &  /         \\

\end{tabular}
\caption{Eligibility criteria applied to the screening and full-text assessment processes.}\label{tab:inex}
\end{table}

%% file: tables/RecordsByDataType.tex
\newpage
\def\arraystretch{1.5}
\begin{sidewaystable}
    \centering
    \begin{tabular}{p{0.6cm} p{0.4cm} p{2.0cm} p{1.3cm} p{3.0cm} p{2.5cm} p{3.8cm} p{4.5cm} p{4.5cm}}
         \tiny Datatype & \tiny Year & \tiny Author & \tiny Field & \tiny Measurement Process & \tiny Timeliness & \tiny Representativeness & \tiny Informativeness & \tiny Consistency \\
         \hline
\raggedright \tiny general data & \tiny 1996 & \tiny \citeauthor{Wang1996} & \tiny non-life-science & \raggedright \tiny accuracy (accuracy), carelessness (objectivity), source credibility (reputation), traceability (traceability), data poisoning 
(objectivity) & \raggedright \tiny age (timeliness) & \raggedright \tiny variety (variety of data and data sources), variety in demographics (completeness), depth of data (depth of data), coverage (completeness) & \raggedright \tiny understandability (interpretability, ease of understanding), conciseness (conciseness), feature importance (value-added (usefulness), task relevance, completeness) & \raggedright \tiny syntactic consistency (representational consistency), plausibility (believeability) \\
\raggedright \tiny general data & \tiny 1997 & \tiny \citeauthor{Redman1996} & \tiny non-life-science & \raggedright \tiny precision (precision), traceability (identifyability) & \raggedright \tiny  & \raggedright \tiny Representativeness (naturalness), granularity (granularity, format precision), coverage (comprehensiveness) & \raggedright \tiny understandability (interpretability, format appropriateness), uniqueness (minimum unnecessary redundancy), feature importance (essentialness, task relevance) & \raggedright \tiny syntactic consistency (structural consistency, representation consistency), semantic consistency (semantic consistency), homogeneity (homogeneity, robustness) \\
\raggedright \tiny general data & \tiny 2000 & \tiny \citeauthor{yoon2000managing} & \tiny non-life-science & \raggedright \tiny accuracy (accuracy, correctness), completeness (completeness), source credibility (reliability) & \raggedright \tiny age (timeliness), currency (currency) & \raggedright \tiny granularity (precision, attribute granularity, detail), coverage (comprehensiveness) & \raggedright \tiny understandability (interpretability, naturalness, clarity, unambiguousness, robustness, presentation appropriateness, media, order), conciseness (conciseness, essentialness), uniqueness (minimally redundant), feature importance (relevance) & \raggedright \tiny Consistency (Consistency) \\
\raggedright \tiny general data & \tiny 2002 & \tiny \citeauthor{Pipino2002} & \tiny non-life-science & \raggedright \tiny accuracy (free-of-error), carelessness (objectivity), completeness (completeness), source credibility (reputation), data poisoning (objectivity) & \raggedright \tiny currency (timeliness) & \raggedright \tiny depth of data (depth of data) & \raggedright \tiny understandability (understandability, interpretability), conciseness (conciseness), feature importance (value-added (usefulness), task relevance) & \raggedright \tiny syntactic consistency (representational consistency), plausibility (believeability) \\
\raggedright \tiny general data & \tiny 2003 & \tiny \citeauthor{Kim2003} & \tiny non-life-science & \raggedright \tiny completeness (missing data) & \raggedright \tiny  & \raggedright \tiny  & \raggedright \tiny  & \raggedright \tiny syntactic consistency (integrity, ambiguous data, Different representations of compound data), compliance (non-standard conforming data), semantic consistency (Inconsistency across multiple tables/file, Different data for the same entity) \\
\raggedright \tiny general data & \tiny 2007 & \tiny \citeauthor{Stvilia2007} & \tiny non-life-science & \raggedright \tiny accuracy (accuracy), precision (precision), source credibility (authority), traceability (verifyability) & \raggedright \tiny age (currency), currency (volatility) & \raggedright \tiny  & \raggedright \tiny Informativeness (Informativeness), understandability (complexity), conciseness (cohesiveness), feature importance (relevance) & \raggedright \tiny syntactic consistency (intrinsic accuracy/validity, intrinsic structural consistency, intrinsic naturalness, intrinsic semantic consistency), compliance (relational semantic consistency, relational structural consistency), semantic consistency (accuracy/validity, relational accuracy) \\
\raggedright \tiny general data & \tiny 2008 & \tiny \citeauthor{iso2500025012} & \tiny non-life-science & \raggedright \tiny completeness (completeness), source credibility (source credibility, authenticity), traceability (traceability) & \raggedright \tiny age (currency) & \raggedright \tiny granularity (precision (no. of digits)) & \raggedright \tiny understandability (understandability), conciseness (efficiency) & \raggedright \tiny Consistency (Consistency), syntactic consistency (syntactic accuracy), compliance (compliance), semantic consistency (semantic accuracy) \\
\raggedright \tiny general data & \tiny 2010 & \tiny \citeauthor{chan2010electronic} & \tiny life science & \raggedright \tiny accuracy (accuracy), completeness (completeness) & \raggedright \tiny currency (timeliness) & \raggedright \tiny granularity (granularity, clinical specificity) & \raggedright \tiny  & \raggedright \tiny semantic consistency (comparability) \\
\raggedright \tiny general data & \tiny 2011 & \tiny \citeauthor{Loshin2011} & \tiny non-life-science & \raggedright \tiny accuracy (value accuracy), precision (precision), completeness (Null validation), source credibility (authoritative sources), traceability (originating data source) & \raggedright \tiny age (age), currency (correction/update promulgation) & \raggedright \tiny coverage (coverage, population density optionality) & \raggedright \tiny uniqueness (Entity uniqueness) & \raggedright \tiny syntactic consistency (syntactic consistency, Name ambiguity), compliance (standards and policies), semantic consistency (semantic consistency, multi-value consistency), plausibility (temporal consistency, reasonableness), distribution drift (temporal reasonability) \\
\raggedright \tiny general data & \tiny 2012 & \tiny \citeauthor{nahm2012data} & \tiny life science & \raggedright \tiny accuracy (accuracy), human-induced error (human-induced error), completeness (completeness) & \raggedright \tiny timeliness (timeliness), currency (currency, volatility) & \raggedright \tiny granularity (granularity, precision (no. of digits)) & \raggedright \tiny understandability (readability/legibility), feature importance (relevance) & \raggedright \tiny  \\
\raggedright \tiny general data & \tiny 2012 & \tiny \citeauthor{Sidi2012} & \tiny non-life-science & \raggedright \tiny accuracy (accuracy, free-of-error, reliability), carelessness (objectivity), completeness (completeness), source credibility (reputation), data poisoning (objectivity) & \raggedright \tiny age (freshness), currency (freshness) & \raggedright \tiny variety in demographics (completeness), depth of data (depth of data), coverage (coverage, completeness) & \raggedright \tiny understandability (understandability, interpretability), conciseness (conciseness), uniqueness (duplication), feature importance (value-added (usefulness), Usability, completeness, relevance) & \raggedright \tiny syntactic consistency (representational consistency, consistency, data integrity fundamentals), semantic consistency (consistency, consistency and synchronization), plausibility (believeability), distribution drift (data decay) \\
\raggedright \tiny general data & \tiny 2012 & \tiny \citeauthor{sebastian2012measuring} & \tiny non-life-science & \raggedright \tiny completeness (completeness) & \raggedright \tiny timeliness (timeliness) & \raggedright \tiny  & \raggedright \tiny  & \raggedright \tiny syntactic consistency (validity), distribution drift ((temporal) consistency) \\
\raggedright \tiny general data & \tiny 2013 & \tiny \citeauthor{dama2013} & \tiny non-life-science & \raggedright \tiny accuracy (accuracy), completeness (completeness) & \raggedright \tiny age (timeliness) & \raggedright \tiny  & \raggedright \tiny uniqueness (uniqueness) & \raggedright \tiny Consistency (Consistency), syntactic consistency (validity, consistency) \\
\raggedright \tiny general data & \tiny 2013 & \tiny \citeauthor{Weiskopf2013} & \tiny life science & \raggedright \tiny accuracy (correctness), carelessness (correctness), completeness (completeness) & \raggedright \tiny currency (currency) & \raggedright \tiny  & \raggedright \tiny  & \raggedright \tiny semantic consistency (concordance), plausibility (plausibility) \\
\end{tabular}
\end{sidewaystable}
\newpage
\def\arraystretch{1.5}
\begin{sidewaystable}
    \centering
    \begin{tabular}{p{0.6cm} p{0.4cm} p{2.0cm} p{1.3cm} p{3.0cm} p{2.5cm} p{3.8cm} p{4.5cm} p{4.5cm}}
         \tiny Datatype & \tiny Year & \tiny Author & \tiny Field & \tiny Measurement Process & \tiny Timeliness & \tiny Representativeness & \tiny Informativeness & \tiny Consistency \\
         \hline
\raggedright \tiny general data & \tiny 2014 & \tiny \citeauthor{Chen2014} & \tiny life science & \raggedright \tiny accuracy (accuracy, reflecting actual sample, reliability), precision (precision), carelessness (objectivity, reflecting actual sample, reliability, under-reporting, illegible handwriting, errors in report form / errors from data entry / calculation errors), completeness (completeness, completeness), expertise (repeatability), data poisoning (objectivity) & \raggedright \tiny timeliness (timeliness), age (periodicity), currency (currency, up-datedness) & \raggedright \tiny Representativeness (Representativeness), granularity (granularity) & \raggedright \tiny understandability (ease of understanding), feature importance (Usability, importance, relevance) & \raggedright \tiny Consistency (Consistency), syntactic consistency (validity, internal consistency, external consistency, non-standardization of vocabulary), compliance (meeting/using data standards), semantic consistency (concordance, comparability) \\
\raggedright \tiny general data & \tiny 2015 & \tiny \citeauthor{Johnson2015} & \tiny life science & \raggedright \tiny Measurement Process (Reliability), accuracy (RelativeCorrectness, RepresentationCorrectness, RelativeCompleteness), carelessness (RepresentationIntegrity), completeness (RepresentationCorrectness, RepresentationComplete, RelativeCompleteness) & \raggedright \tiny currency (RepresentationCurrent, DatasetCurrent, TaskCurrency) & \raggedright \tiny variety in demographics (sufficiency), coverage (DomainCoverage, TaskCoverage) & \raggedright \tiny informative missingness (DomainComplete), feature importance (relevance) & \raggedright \tiny syntactic consistency (RepresentationConsistency), compliance (CodingConsistency), semantic consistency (DomainConsistency) \\
\raggedright \tiny general data & \tiny 2016 & \tiny \citeauthor{Kahn2016} & \tiny life science & \raggedright \tiny completeness (completeness) & \raggedright \tiny  & \raggedright \tiny  & \raggedright \tiny uniqueness (uniqueness) & \raggedright \tiny syntactic consistency (value conformance, relational conformance, computational conformance), plausibility (plausibility) \\
\raggedright \tiny general data & \tiny 2018 & \tiny \citeauthor{che2018recurrent} & \tiny life science & \raggedright \tiny  & \raggedright \tiny  & \raggedright \tiny  & \raggedright \tiny informative missingness (informative missingness) & \raggedright \tiny  \\
\raggedright \tiny general data & \tiny 2019 & \tiny \citeauthor{Bloland2019} & \tiny life science & \raggedright \tiny accuracy (Trueness), carelessness (integrity), completeness (completeness), data poisoning (integrity) & \raggedright \tiny age (timeliness) & \raggedright \tiny coverage (completeness) & \raggedright \tiny conciseness (efficiency), feature importance (relevance) & \raggedright \tiny syntactic consistency (consistency), semantic consistency (concurrence) \\
\raggedright \tiny general data & \tiny 2019 & \tiny \citeauthor{Vanbrabant2019} & \tiny life science & \raggedright \tiny carelessness (Inexactness of timestamps), completeness (missing values) & \raggedright \tiny  & \raggedright \tiny granularity (imprecise data), coverage (missing entities) & \raggedright \tiny  & \raggedright \tiny syntactic consistency (typing mistake, outside domain range, inconsistent formatting, abbreviations), semantic consistency (violation of mutual dependency) \\
\raggedright \tiny general data & \tiny 2020 & \tiny \citeauthor{Bian2020} & \tiny life science & \raggedright \tiny accuracy (accuracy) & \raggedright \tiny currency (currency) & \raggedright \tiny  & \raggedright \tiny uniqueness (uniqueness) & 
\raggedright \tiny syntactic consistency (value conformance, relational conformance), semantic consistency (concordance, comparability), plausibility (plausibility), distribution drift (temporal plausbility) \\
\raggedright \tiny general data & \tiny 2021 & \tiny \citeauthor{Schmidt2021} & \tiny life science & \raggedright \tiny accuracy (accuracy, disagreement with gold standard), carelessness (inter-class/intra-class reliability), outliers (outliers, unexpected distributions, unexpected associations, unexpected data elements, unexpected data records), completeness (completeness, completeness, crude missingness), expertise (disagreement of repeated mesurements) & \raggedright \tiny  & \raggedright \tiny  & \raggedright \tiny understandability (readability/legibility), uniqueness (duplication), informative missingness (qualified missingness, non response rate, refusal rate, dropout rate, missing due to specified reason) & \raggedright \tiny Consistency (Consistency), syntactic consistency (structural data set error, value format error, relational data set error, range and value violations, non-standardization of vocabulary, data type mismatch, inhomogeneous value formats, inadmissible precision), compliance (integrity), semantic consistency (semantic consistency, accuracy/validity, contradictions, data record mismatch, data element mismatch) \\
\raggedright \tiny general data & \tiny 2023 & \tiny \citeauthor{Syed2023} & \tiny life science & \raggedright \tiny accuracy (accuracy, correctness, veracity), noisy labels (accurate diagnostic data), carelessness (integrity), completeness (missing data, fragmentation), data poisoning (integrity) & \raggedright \tiny age (timeliness) & \raggedright \tiny Representativeness (Representativeness), granularity (granularity), coverage (completeness) & \raggedright \tiny uniqueness (uniqueness), feature importance (completeness, relevance) & \raggedright \tiny syntactic consistency (representational consistency, validity, structuredness), compliance (integrity, standardization,), semantic consistency (semantic consistency, concordance), plausibility (plausibility), homogeneity (data variability), distribution drift (temporal variability) \\
\raggedright \tiny general data & \tiny 2023 & \tiny \citeauthor{Mashoufi2023} & \tiny life science & \raggedright \tiny accuracy (accuracy, correctness, validity), noisy labels (data entry error), carelessness (objectivity), completeness (completeness), source credibility (source credibility), data poisoning (objectivity) & \raggedright \tiny age (currency), currency (timeliness, temporal consistency) & \raggedright \tiny granularity (granularity) & \raggedright \tiny feature importance (relevancy) & \raggedright \tiny Consistency (Consistency), syntactic consistency (conformance), semantic consistency (semantic plausibility) \\
\end{tabular}
\begin{tabular}{cc}
     & \\
     & \\
     & \\
\end{tabular}
\end{sidewaystable}
\newpage
\def\arraystretch{1.5}
\begin{sidewaystable}
    \centering
    \begin{tabular}{p{0.6cm} p{0.4cm} p{2.0cm} p{1.3cm} p{3.0cm} p{2.5cm} p{3.8cm} p{4.5cm} p{4.5cm}}
         \tiny Datatype & \tiny Year & \tiny Author & \tiny Field & \tiny Measurement Process & \tiny Timeliness & \tiny Representativeness & \tiny Informativeness & \tiny Consistency \\
         \hline
\raggedright \tiny general data & \tiny 2023 & \tiny \citeauthor{Lewis2023} & \tiny life science & \raggedright \tiny accuracy (correctness), carelessness (correctness), completeness (completeness) & \raggedright \tiny currency (currency) & \raggedright \tiny  & \raggedright \tiny informative missingness (bias) & \raggedright \tiny semantic consistency (concordance), plausibility (plausibility) \\
\raggedright \tiny general data & \tiny 2023 & \tiny \citeauthor{Liu2023} & \tiny life science & \raggedright \tiny accuracy (RelativeCorrectness), carelessness (correctness, misunderstanding of what an entry means, errors in report form / errors 
from data entry / calculation errors) & \raggedright \tiny age (timeliness), currency (currency) & \raggedright \tiny dataset size (completeness), coverage (completeness (coverage of baseline features or data required for a particular disease)) & 
\raggedright \tiny understandability (usability, ease of understanding), uniqueness (duplication), feature importance (relevance) & \raggedright \tiny syntactic consistency (syntactic consistency, consistency through integrity constraints), distribution drift ((temporal) consistency) \\
\raggedright \tiny general data & \tiny 2023 & \tiny \citeauthor{Xu2023} & \tiny non-life-science & \raggedright \tiny noisy labels (noisy labels), source credibility (reliability) & \raggedright \tiny  & \raggedright \tiny dataset size (dataset size) & \raggedright \tiny  & \raggedright \tiny  \\
\raggedright \tiny general data & \tiny 2023 & \tiny \citeauthor{Tahar2023} & \tiny life science & \raggedright \tiny completeness (completeness, value completeness, item completeness) & \raggedright \tiny  & \raggedright \tiny  & \raggedright \tiny conciseness (semantic uniqueness), uniqueness (syntactic uniqueness) & \raggedright \tiny syntactic consistency (range plausibility), semantic consistency (semantic plausibility, concordance), plausibility (plausibility) \\
\raggedright \tiny big data & \tiny 2015 & \tiny \citeauthor{Batini2015} & \tiny non-life-science & \raggedright \tiny accuracy (accuracy, semantic accuracy, source accuracy, reliability), precision (precision, accuracy deviation), noisy labels (correctness of classification), completeness (completeness, value completeness, tuple completeness, left censored, right censored, sparsity), source credibility (reputation, trustworthiness), expertise (believeability), traceability (verifyability) & \raggedright \tiny age (timeliness), currency (currency, temporal validity, temporal consistency) & \raggedright \tiny Representativeness (Representativeness) & \raggedright \tiny understandability (understandability, readability/legibility), conciseness (conciseness, closer-to-text base comprehension, Closer to situation model level comprehension, representation conc.), uniqueness (spatial redundancy, temporal redudancy), feature importance (significance, pertinence, relevance, selective) & \raggedright \tiny Consistency (Consistency), syntactic consistency (syntactic accuracy, consistency through integrity constraints, consistency through edits, format consistency, domain consistency, lexical consistency, numerical consistency, range frequency, change frequency), logical consistency (logical consistency), semantic consistency (relative error consistency, coherence), plausibility (temporal consistency) \\
\raggedright \tiny big data & \tiny 2015 & \tiny \citeauthor{Cai2015} & \tiny non-life-science & \raggedright \tiny accuracy (accuracy), source credibility (source credibility) & \raggedright \tiny age (timeliness), currency (timeliness) & \raggedright \tiny coverage (completeness) & \raggedright \tiny understandability (readability), feature importance (fitness) & \raggedright \tiny syntactic consistency (consistency, integrity), semantic consistency (consistency, integrity) \\
\raggedright \tiny big data & \tiny 2016 & \tiny \citeauthor{Gao2016} & \tiny non-life-science & \raggedright \tiny accuracy (accuracy, correctness), carelessness (correctness), completeness (completeness), source credibility (accountability) & \raggedright \tiny timeliness (timeliness), currency (currency) & \raggedright \tiny  & \raggedright \tiny feature importance (Usability) & \raggedright \tiny syntactic consistency (format consistency) \\
\raggedright \tiny big data & \tiny 2017 & \tiny \citeauthor{gudivada2017data} & \tiny non-life-science & \raggedright \tiny carelessness (gender bias), completeness (missing at random (MAR), missing completely at random (MCAR)), data poisoning (gender bias) & \raggedright \tiny currency (currency) & \raggedright \tiny variety in demographics (heterogeneity), variety of data sources (heterogeneity) & \raggedright \tiny conciseness (dimensionality reduction), uniqueness (duplication, consistency), informative missingness (missing not at random (MNAR)), feature importance (feature selection, feature extraction) & \raggedright \tiny syntactic consistency (syntactic accuracy), compliance (specifications, integrity), semantic consistency (semantic accuracy) \\
\raggedright \tiny big data & \tiny 2018 & \tiny \citeauthor{Juddoo2018} & \tiny life science & \raggedright \tiny accuracy (accuracy), carelessness (reliability), completeness (completeness), expertise (confidence) & \raggedright \tiny  & \raggedright \tiny  & \raggedright \tiny feature importance (usefulness) & \raggedright \tiny syntactic consistency (consistency), semantic consistency (consistency), plausibility (validity) \\
\raggedright \tiny big data & \tiny 2020 & \tiny \citeauthor{Eder2020} & \tiny life science & \raggedright \tiny accuracy (accuracy), completeness (completeness), source credibility (Reliability) & \raggedright \tiny age (timeliness) & \raggedright \tiny granularity (precision (no. of digits)) & \raggedright \tiny feature importance (reliability) & \raggedright \tiny compliance (compliance), semantic consistency (semantic consistency) \\
\raggedright \tiny big data & \tiny 2020 & \tiny \citeauthor{Ramasamy2020} & \tiny non-life-science & \raggedright \tiny source credibility (source credibility, trustworthiness), traceability (Pedigree/Lineage) & \raggedright \tiny  & \raggedright \tiny dataset size (volume) & \raggedright \tiny redundancy (redundancy) & \raggedright \tiny semantic consistency (cohesion) \\
\raggedright \tiny machine learning & \tiny 1993 & \tiny \citeauthor{Bansal1993} & \tiny non-life-science & \raggedright \tiny carelessness (typing error (by white noise), measurement of subjective data (e.g., about consumer sentiment) -> as white noise, inaccurate estimations (e.g., from forecasts)) & \raggedright \tiny  & \raggedright \tiny  & \raggedright \tiny  & \raggedright \tiny  \\
\raggedright \tiny machine learning & \tiny 2004 & \tiny \citeauthor{Michel2000} & \tiny life science & \raggedright \tiny  & \raggedright \tiny  & \raggedright \tiny dataset size (quantity), coverage (completeness) & \raggedright \tiny  & \raggedright \tiny  \\
\end{tabular}
\end{sidewaystable}
\newpage
\def\arraystretch{1.5}
\begin{sidewaystable}
    \centering
    \begin{tabular}{p{0.6cm} p{0.4cm} p{2.0cm} p{1.3cm} p{3.0cm} p{2.5cm} p{3.8cm} p{4.5cm} p{4.5cm}}
         \tiny Datatype & \tiny Year & \tiny Author & \tiny Field & \tiny Measurement Process & \tiny Timeliness & \tiny Representativeness & \tiny Informativeness & \tiny Consistency \\
         \hline
\raggedright \tiny machine learning & \tiny 2011 & \tiny \citeauthor{Blake2011} & \tiny non-life-science & \raggedright \tiny accuracy (accuracy), completeness (completeness) & \raggedright \tiny age (timeliness) & \raggedright \tiny  & \raggedright \tiny  & \raggedright \tiny syntactic consistency (consistency) \\
\raggedright \tiny machine learning & \tiny 2013 & \tiny \citeauthor{twala2013impact} & \tiny non-life-science & \raggedright \tiny precision (noisy attributes), noisy labels (noisy targets), carelessness (noisy attributes), outliers (outliers) & 
\raggedright \tiny  & \raggedright \tiny  & \raggedright \tiny  & \raggedright \tiny  \\
\raggedright \tiny machine learning & \tiny 2015 & \tiny \citeauthor{Sukhbaatar2014} & \tiny non-life-science & \raggedright \tiny noisy labels (noisy labels, label flips, outliers) & \raggedright \tiny  & \raggedright \tiny  & \raggedright \tiny 
 & \raggedright \tiny  \\
\raggedright \tiny machine learning & \tiny 2015 & \tiny \citeauthor{masko2015impact} & \tiny non-life-science & \raggedright \tiny  & \raggedright \tiny  & \raggedright \tiny target class balance (imbalance) & \raggedright \tiny  & \raggedright \tiny  \\
\raggedright \tiny machine learning & \tiny 2016 & \tiny \citeauthor{dodge2016understanding} & \tiny non-life-science & \raggedright \tiny accuracy (compression/noise/blur/contrast) & \raggedright \tiny  & \raggedright \tiny  & \raggedright \tiny 
 & \raggedright \tiny  \\
\raggedright \tiny machine learning & \tiny 2016 & \tiny \citeauthor{Karahan2016} & \tiny non-life-science & \raggedright \tiny accuracy (contrast, noise, blur, occlusion, and color degredation) & \raggedright \tiny  & \raggedright \tiny  & \raggedright \tiny  & \raggedright \tiny  \\
\raggedright \tiny machine learning & \tiny 2017 & \tiny \citeauthor{sun2017revisiting} & \tiny non-life-science & \raggedright \tiny noisy labels (noisy labels) & \raggedright \tiny  & \raggedright \tiny dataset size (dataset size), coverage (number of classes) & \raggedright \tiny  & \raggedright \tiny  \\
\raggedright \tiny machine learning & \tiny 2018 & \tiny \citeauthor{Rolnick2017} & \tiny non-life-science & \raggedright \tiny noisy labels (noisy labels, uniform label noise, structured label noise) & \raggedright \tiny  & \raggedright \tiny dataset size (dataset size) & \raggedright \tiny  & \raggedright \tiny  \\
\raggedright \tiny machine learning & \tiny 2018 & \tiny \citeauthor{Buda2018} & \tiny non-life-science & \raggedright \tiny  & \raggedright \tiny  & \raggedright \tiny target class balance (target class balance) & \raggedright \tiny  & \raggedright \tiny  \\
\raggedright \tiny machine learning & \tiny 2019 & \tiny \citeauthor{He2019} & \tiny non-life-science & \raggedright \tiny noisy labels (quality of label), data poisoning (dataset contamination) & \raggedright \tiny  & \raggedright \tiny dataset size (dataset size), target class balance (dataset equilibrium) & \raggedright \tiny  & \raggedright \tiny  \\
\raggedright \tiny machine learning & \tiny 2019 & \tiny \citeauthor{ovadia2019can} & \tiny non-life-science & \raggedright \tiny  & \raggedright \tiny  & \raggedright \tiny  & \raggedright \tiny  & \raggedright \tiny distribution drift (distribution drift) \\
\raggedright \tiny machine learning & \tiny 2020 & \tiny \citeauthor{Cao2020} & \tiny non-life-science & \raggedright \tiny  & \raggedright \tiny  & \raggedright \tiny  & \raggedright \tiny uniqueness (rank) & \raggedright \tiny  \\
\raggedright \tiny machine learning & \tiny 2021 & \tiny \citeauthor{Hong2021} & \tiny life science & \raggedright \tiny accuracy (adding white noise) & \raggedright \tiny  & \raggedright \tiny dataset size (dataset size) & \raggedright \tiny  & \raggedright \tiny  \\
\raggedright \tiny machine learning & \tiny 2021 & \tiny \citeauthor{Benedick2021} & \tiny non-life-science & \raggedright \tiny accuracy (swapping perturbation), completeness (dropping perturbation) & \raggedright \tiny  & \raggedright \tiny  & \raggedright \tiny  & \raggedright \tiny  \\
\raggedright \tiny machine learning & \tiny 2021 & \tiny \citeauthor{Wesemeyer2021} & \tiny life science & \raggedright \tiny noisy labels (annotation quality) & \raggedright \tiny  & \raggedright \tiny dataset size (quantity) & \raggedright \tiny  & \raggedright \tiny  \\
\raggedright \tiny machine learning & \tiny 2021 & \tiny \citeauthor{Qi2021} & \tiny non-life-science & \raggedright \tiny completeness (missing data) & \raggedright \tiny  & \raggedright \tiny  & \raggedright \tiny  & \raggedright \tiny semantic 
consistency (violations of functional dependencies, conflicting data) \\
\raggedright \tiny machine learning & \tiny 2021 & \tiny \citeauthor{Li2021} & \tiny non-life-science & \raggedright \tiny  & \raggedright \tiny  & \raggedright \tiny dataset size (quantity) & \raggedright \tiny feature importance (information value) & \raggedright \tiny  \\
\raggedright \tiny machine learning & \tiny 2021 & \tiny \citeauthor{LiYang2021} & \tiny non-life-science & \raggedright \tiny  & \raggedright \tiny  & \raggedright \tiny target class balance (target class balance) & \raggedright \tiny redundancy 
(redundancy), feature importance (informative sample) & \raggedright \tiny  \\
\raggedright \tiny machine learning & \tiny 2022 & \tiny \citeauthor{Fan2022} & \tiny life science & \raggedright \tiny accuracy (adding random errors) & \raggedright \tiny  & \raggedright \tiny dataset size (quantity) & \raggedright \tiny  & \raggedright \tiny  \\
\raggedright \tiny machine learning & \tiny 2022 & \tiny \citeauthor{Jouseau2022} & \tiny non-life-science & \raggedright \tiny outliers (outliers), completeness (missing values) & \raggedright \tiny  & \raggedright \tiny  & \raggedright \tiny uniqueness (exact and partial dublicates), informative missingness (missing values) & \raggedright \tiny syntactic consistency (domain value violations) \\
\raggedright \tiny machine learning & \tiny 2022 & \tiny \citeauthor{Derry2022} & \tiny life science & \raggedright \tiny  & \raggedright \tiny  & \raggedright \tiny dataset size (more training data), granularity (downsampling) & \raggedright \tiny feature importance (use of different experimental methods) & \raggedright \tiny  \\
\end{tabular}
\begin{tabular}{cc}
     & \\
     & \\
\end{tabular}
\end{sidewaystable}
\newpage
\def\arraystretch{1.5}
\begin{sidewaystable}
    \centering
    \begin{tabular}{p{0.6cm} p{0.4cm} p{2.0cm} p{1.3cm} p{3.0cm} p{2.5cm} p{3.8cm} p{4.5cm} p{4.5cm}}
         \tiny Datatype & \tiny Year & \tiny Author & \tiny Field & \tiny Measurement Process & \tiny Timeliness & \tiny Representativeness & \tiny Informativeness & \tiny Consistency \\
         \hline
\raggedright \tiny machine learning & \tiny 2023 & \tiny \citeauthor{Ranjan2023} & \tiny non-life-science & \raggedright \tiny accuracy (lighting condition), precision (augmentation (mixup/brightness/exposure/saturation/noise/blur/cutout/mosaic)) 
& \raggedright \tiny  & \raggedright \tiny dataset size (dataset size, augmentation) & \raggedright \tiny  & \raggedright \tiny  \\
\raggedright \tiny machine learning & \tiny 2023 & \tiny \citeauthor{Whang2023} & \tiny non-life-science & \raggedright \tiny precision (augmentation (mixup/brightness/exposure/saturation/noise/blur/cutout/mosaic), noisy features (image blur/color noise)), noisy labels (noisy labels), carelessness (racial bias, fairness), completeness (missing values, missing labels), data poisoning (data poisoning, label flipping, racial bias, fairness) & \raggedright \tiny  & \raggedright \tiny  & \raggedright \tiny informative missingness (informative missingness) & \raggedright \tiny  \\
\raggedright \tiny machine learning & \tiny 2023 & \tiny \citeauthor{shimizu2022effect} & \tiny life science & \raggedright \tiny noisy labels (label redudancy) & \raggedright \tiny  & \raggedright \tiny dataset size (budget for crowdsourcing) & \raggedright \tiny  & \raggedright \tiny  \\
\raggedright \tiny machine learning & \tiny 2023 & \tiny \citeauthor{Pan2023} & \tiny non-life-science & \raggedright \tiny  & \raggedright \tiny  & \raggedright \tiny target class balance (target class balance) & \raggedright \tiny  & \raggedright \tiny homogeneity (interclass ambiguity, interclass ambiguity, out of distribution) \\
    \end{tabular}
\caption{List of all publications in the literature corpus including datatype (general data, big data or ML data), publication year, authors, and field (life-science vs. not-life-science). Additionally, we list the mentioned data quality dimensions and subdimensions per paper, categorised into the clusters of the METRIC-framework (Measurement Process, Timeliness, Representativeness, Informativeness, Consistency). The originally mentioned data quality terms (displayed in brackets) are mapped to the corresponding dimension or subdimension of the METRIC-framework and displayed in the form: \q{METRIC-dimension (terms used in publication)}, e.g., \q{dataset size (quantity)}.}
\begin{tabular}{cc}
     & \\
     & \\
     & \\
     & \\
     & \\
     & \\
     & \\
     & \\
     & \\
     & \\
     & \\
     & \\
     & \\
     & \\
     & \\
     & \\
     & \\
     & \\
     & \\
     & \\
     & \\
\end{tabular}
\end{sidewaystable}

%% file: tables/DQD_to_papers.tex
\begin{table}[ht!]
\begin{tabular}{p{10.7cm} p{5.3cm} p{3.7cm} p{3.5cm}}
\bf Measurement Process \normalfont \cite{Johnson2015}&  & \ & \\
\end{tabular}
\begin{tabular}{| p{3.7cm} | p{3.3cm} | p{3.7cm} |  p{3.5cm} |}
\hline
Dimension & Subdimension & Definition & References \\
\hline
\raggedright device error & \ & \footnotesize The extent to which data values originating from a sensor are accurate and precise. &\footnotesize \\
\cdashline{2-4}[0.5pt/2.5pt]
\ & accuracy & \footnotesize The extent of closeness of data values to real values. (From \cite{dama2020}) &\footnotesize \cite{Cai2015,Syed2023,Johnson2015,Batini2015,Gao2016,Chen2014,Schmidt2021,Mashoufi2023,yoon2000managing,nahm2012data,Ranjan2023,Hong2021,Karahan2016,dama2013,Liu2023,Bloland2019,Pipino2002,Fan2022,Eder2020,Loshin2011,Sidi2012,Blake2011,chan2010electronic,dodge2016understanding,Wang1996,Bian2020,Lewis2023,Benedick2021,Juddoo2018,Stvilia2007,Weiskopf2013} \\
\cdashline{2-4}[0.5pt/2.5pt]
\ & precision & \footnotesize The extent to which the error in data values spreads around zero (in statistics). (From \cite{dama2020}) &\footnotesize \cite{Batini2015,Whang2023,Chen2014,Redman1996,Stvilia2007,Loshin2011,Ranjan2023,twala2013impact} \\
\cdashline{1-4}
\raggedright human-induced error & \ & \footnotesize The extent to which manual inputs unintentionally influence the data in a wrong way. &\footnotesize \cite{nahm2012data} \\
\cdashline{2-4}[0.5pt/2.5pt]
\ & noisy labels & \footnotesize The extent to which labels provided by humans are accurate and precise. &\footnotesize \cite{Sukhbaatar2014,Syed2023,Whang2023,Batini2015,Wesemeyer2021,Rolnick2017,sun2017revisiting,Mashoufi2023,shimizu2022effect,Xu2023,twala2013impact,He2019} \\
\cdashline{2-4}[0.5pt/2.5pt]
\ & carelessness & \footnotesize The extent to which negligence or unintended human errors cause faulty data. &\footnotesize \cite{Johnson2015,Syed2023,Whang2023,Gao2016,Mashoufi2023,Chen2014,Schmidt2021,Vanbrabant2019,Liu2023,Bloland2019,Pipino2002,Sidi2012,Bansal1993,Wang1996,Lewis2023,gudivada2017data,Juddoo2018,twala2013impact,Weiskopf2013} \\
\cdashline{2-4}[0.5pt/2.5pt]
\ & outliers & \footnotesize The extent to which data values exist that do not match the expected distribution. &\footnotesize \cite{Jouseau2022,Schmidt2021,twala2013impact} \\
\cdashline{1-4}
\raggedright completeness & \ & \footnotesize The extent to which data values are present. (From \cite{dama2020}) &\footnotesize \cite{Qi2021,Johnson2015,Syed2023,Batini2015,Whang2023,Gao2016,Chen2014,Schmidt2021,Mashoufi2023,nahm2012data,yoon2000managing,Vanbrabant2019,Jouseau2022,sebastian2012measuring,iso2500025012,dama2013,Kim2003,Bloland2019,Pipino2002,Eder2020,Loshin2011,Sidi2012,Blake2011,chan2010electronic,Lewis2023,gudivada2017data,Benedick2021,Juddoo2018,Tahar2023,Kahn2016,Weiskopf2013} \\
\cdashline{1-4}
\raggedright source credibility & \ & \footnotesize The extent to which data values are regarded as true and believable in terms of their source. (Based on \cite{dama2020}) &\footnotesize \cite{Xu2023,Sidi2012,Cai2015,iso2500025012,Batini2015,Gao2016,Ramasamy2020,Mashoufi2023,Pipino2002,Eder2020,Stvilia2007,yoon2000managing,Loshin2011,Wang1996} \\
\cdashline{2-4}[0.5pt/2.5pt]
\ & expertise & \footnotesize The extent of knowledge and experience with which the measurement was performed and the data was processed. &\footnotesize \cite{Chen2014,Batini2015,Schmidt2021,Juddoo2018} \\
\cdashline{2-4}[0.5pt/2.5pt]
\ & traceability & \footnotesize The extent to which data lineage is available. (From \cite{dama2020}) &\footnotesize \cite{iso2500025012,Batini2015,Ramasamy2020,Redman1996,Stvilia2007,Loshin2011,Wang1996} \\
\cdashline{2-4}[0.5pt/2.5pt]
\ & data poisoning & \footnotesize The extent to which data values are intentionally falsified. &\footnotesize \cite{Syed2023,Wang1996,gudivada2017data,Whang2023,Bloland2019,Pipino2002,Chen2014,Mashoufi2023,Sidi2012,He2019} \\
\hline
\end{tabular}
\caption{Details for the cluster \textit{Measurement Process} of the METRIC-framework. The hierarchical order of all dimensions and subdimensions included in the cluster is displayed in the first two columns. The third column provides a definition for each of these terms. The fourth column provides all references of the literature corpus connected to the corresponding term. A provided reference indicates that the dimension, subdimension or a different term with the same intended meaning (see \nameref{sec:dataextraction} for clarification) is discussed in that cited article. References on the cluster level indicate that the term is used as a broader concept.}
\label{tab:dqd_to_papers1}
\end{table}
\begin{table}[ht!]
\begin{tabular}{p{10.7cm} p{5.3cm} p{3.7cm} p{3.5cm}}
\bf Timeliness& \ & \ &\footnotesize \\
\end{tabular}
\begin{tabular}{| p{3.7cm} | p{3.3cm} | p{3.7cm} |  p{3.5cm} |}
\hline
Dimension & Subdimension & Definition & References \\
\hline
\raggedright timeliness & \ & \footnotesize The extent to which the \textit{age} and the \textit{currency} of data values are appropriate for the task at hand. (Based on \cite{Wang1996}) &\footnotesize \cite{Gao2016,Chen2014,sebastian2012measuring,nahm2012data} \\  
\cdashline{2-4}[0.5pt/2.5pt]
\ & age & \footnotesize The extent to which the age of data values is appropriate for the task at hand. (Based on \cite{Wang1996}) &\footnotesize \cite{Blake2011,Sidi2012,Cai2015,iso2500025012,Syed2023,Batini2015,Liu2023,dama2013,Bloland2019,Mashoufi2023,Chen2014,Eder2020,Stvilia2007,yoon2000managing,Loshin2011,Wang1996} \\
\cdashline{2-4}[0.5pt/2.5pt]
\ & currency & \footnotesize The extent to which data values are up to date. (From \cite{dama2020}) &\footnotesize \cite{yoon2000managing,Lewis2023,chan2010electronic,Cai2015,Johnson2015,gudivada2017data,Batini2015,Gao2016,Liu2023,Chen2014,Pipino2002,Mashoufi2023,Stvilia2007,nahm2012data,Loshin2011,Sidi2012,Bian2020,Weiskopf2013} \\
\hline
\end{tabular}
\caption{Details for the cluster \textit{Timeliness} of the METRIC-framework. The hierarchical order of all dimensions and subdimensions included in the cluster is displayed in the first two columns. The third column provides a definition for each of these terms. The fourth column provides all references of the literature corpus connected to the corresponding term. A provided reference indicates that the dimension, subdimension or a different term with the same intended meaning (see \nameref{sec:dataextraction} for clarification) is discussed in that cited article. References on the cluster level indicate that the term is used as a broader concept.}
\label{tab:dqd_to_papers2}
\begin{tabular}{p{10.7cm} p{5.3cm} p{3.7cm} p{3.5cm}}
\bf Representativeness \normalfont \cite{Chen2014,Batini2015,Redman1996,Syed2023}&  & \ & \\
\end{tabular}
\begin{tabular}{| p{3.7cm} | p{3.3cm} | p{3.7cm} |  p{3.5cm} |}
\hline
Dimension & Subdimension & Definition & References \\
\hline
\raggedright variety & \ & \footnotesize The extent to which the dataset is diverse. &\footnotesize \cite{Wang1996} \\
\cdashline{2-4}[0.5pt/2.5pt]
\ & variety in demographics & \footnotesize The extent to which measured subjects are diverse. &\footnotesize \cite{gudivada2017data,Wang1996,Sidi2012,Johnson2015} \\
\cdashline{2-4}[0.5pt/2.5pt]
\ & variety of data sources & \footnotesize The extent to which data are available from different sources. (Based on \cite{dama2020}) &\footnotesize \cite{gudivada2017data} \\
\cdashline{1-4}
\raggedright depth of data & \ & \footnotesize The extent to which the volume of the dataset is appropriate. (Based on \cite{Wang1996}) &\footnotesize \cite{Sidi2012,Wang1996,Pipino2002} \\
\cdashline{2-4}[0.5pt/2.5pt]
\ & dataset size & \footnotesize The extent to which the quantity of records is appropriate. &\footnotesize \cite{Michel2000,Hong2021,Derry2022,Wesemeyer2021,Rolnick2017,Liu2023,Ramasamy2020,sun2017revisiting,Fan2022,Li2021,shimizu2022effect,Xu2023,Ranjan2023,He2019} \\
\cdashline{2-4}[0.5pt/2.5pt]
\ & granularity & \footnotesize The extent to which the level of detail of the dataset and data values is appropriate. &\footnotesize \cite{yoon2000managing,chan2010electronic,iso2500025012,Syed2023,Derry2022,Chen2014,Mashoufi2023,Redman1996,Eder2020,nahm2012data,Vanbrabant2019} \\
\cdashline{2-4}[0.5pt/2.5pt]
\ & coverage & \footnotesize The extent to which relevant subsets of the dataset satisfy the dimensions \textit{variety} and \textit{target class balance}. &\footnotesize \cite{Cai2015,Syed2023,Wang1996,Johnson2015,Liu2023,Bloland2019,sun2017revisiting,Redman1996,yoon2000managing,Loshin2011,Sidi2012,Vanbrabant2019,Michel2000} \\
\cdashline{1-4}
\raggedright target class balance & \ & \footnotesize The extent to which the classes of a target variable have similar size. &\footnotesize \cite{Buda2018,LiYang2021,masko2015impact,Pan2023,He2019} \\
\hline
\end{tabular}
\caption{Details for the cluster \textit{Representativeness} of the METRIC-framework. The hierarchical order of all dimensions and subdimensions included in the cluster is displayed in the first two columns. The third column provides a definition for each of these terms. The fourth column provides all references of the literature corpus connected to the corresponding term. A provided reference indicates that the dimension, subdimension or a different term with the same intended meaning (see \nameref{sec:dataextraction} for clarification) is discussed in that cited article. References on the cluster level indicate that the term is used as a broader concept.}
\label{tab:dqd_to_papers3}
\end{table}
\begin{table}[ht!]
\begin{tabular}{p{10.7cm} p{5.3cm} p{3.7cm} p{3.5cm}}
\bf Informativeness \normalfont \cite{Stvilia2007}&  & \ & \\
\end{tabular}
\begin{tabular}{| p{3.7cm} | p{3.3cm} | p{3.7cm} |  p{3.5cm} |}
\hline
Dimension & Subdimension & Definition & References \\
\hline
\raggedright understandability & \ & \footnotesize The extent to which data values are clear without ambiguity and easily comprehended. (From \cite{Wang1996}) &\footnotesize \cite{iso2500025012,Cai2015,Wang1996,nahm2012data,Batini2015,Liu2023,Pipino2002,Chen2014,Redman1996,Schmidt2021,yoon2000managing,Stvilia2007,Sidi2012} \\
\cdashline{1-4}
\raggedright redundancy & \ & \footnotesize The extent to which logically identical data are stored more than once. (From \cite{dama2020}) &\footnotesize \cite{Ramasamy2020,LiYang2021} \\
\cdashline{2-4}[0.5pt/2.5pt]
\ & conciseness & \footnotesize The extent to which data values, attributes and records are compactly represented. (Based on \cite{Wang1996}) &\footnotesize \cite{iso2500025012,Wang1996,gudivada2017data,Batini2015,Bloland2019,Pipino2002,Stvilia2007,yoon2000managing,Sidi2012,Tahar2023} \\
\cdashline{2-4}[0.5pt/2.5pt]
\ & uniqueness & \footnotesize The extent to which records occur only once in a data file. (From \cite{dama2020}) &\footnotesize \cite{Jouseau2022,Sidi2012,Syed2023,gudivada2017data,Batini2015,Cao2020,dama2013,Liu2023,Schmidt2021,Redman1996,yoon2000managing,Loshin2011,Kahn2016,Bian2020,Tahar2023} \\
\cdashline{1-4}
\raggedright informative missingness & \ & \footnotesize The extent to which missing data values provide useful information. &\footnotesize \cite{che2018recurrent,Lewis2023,Jouseau2022,Johnson2015,gudivada2017data,Whang2023,Schmidt2021} \\
\cdashline{1-4}
\raggedright feature importance & \ & \footnotesize The extent to which the attributes are beneficial and provide advantages from their use. (Based on \textit{usefulness} from \cite{Wang1996}) &\footnotesize \cite{Cai2015,Syed2023,Johnson2015,Batini2015,LiYang2021,Gao2016,Chen2014,Mashoufi2023,yoon2000managing,nahm2012data,Liu2023,Bloland2019,Pipino2002,Redman1996,Eder2020,Sidi2012,Wang1996,gudivada2017data,Derry2022,Juddoo2018,Li2021,Stvilia2007} \\
\hline
\end{tabular}
\caption{Details for the cluster \textit{Informativeness} of the METRIC-framework. The hierarchical order of all dimensions and subdimensions included in the cluster is displayed in the first two columns. The third column provides a definition for each of these terms. The fourth column provides all references of the literature corpus connected to the corresponding term. A provided reference indicates that the dimension, subdimension or a different term with the same intended meaning (see \nameref{sec:dataextraction} for clarification) is discussed in that cited article. References on the cluster level indicate that the term is used as a broader concept.}
\label{tab:dqd_to_papers4}
\begin{tabular}{p{10.7cm} p{5.3cm} p{3.7cm} p{3.5cm}}
\bf Consistency \normalfont \cite{iso2500025012,Batini2015,dama2013,Chen2014,Schmidt2021,Mashoufi2023,yoon2000managing}&  & \ & \\
\end{tabular}
\begin{tabular}{| p{3.7cm} | p{3.3cm} | p{3.7cm} |  p{3.5cm} |}
\hline
Dimension & Subdimension & Definition & References \\
\hline
\raggedright rule-based consistency & \ & \footnotesize The extent to which data values comply with a rule. (\textit{Validity} from \cite{dama2020}) &\footnotesize \\
\cdashline{2-4}[0.5pt/2.5pt]
\ & syntactic consistency & \footnotesize The extent to which data values comply to a dictionary and are always presented in the same format. (Based on \cite{Wang1996}) &\footnotesize \cite{Cai2015,Syed2023,Johnson2015,Batini2015,Gao2016,Schmidt2021,Chen2014,Mashoufi2023,Vanbrabant2019,Jouseau2022,sebastian2012measuring,iso2500025012,Liu2023,dama2013,Bloland2019,Pipino2002,Kim2003,Redman1996,Loshin2011,Sidi2012,Blake2011,Wang1996,Bian2020,gudivada2017data,Juddoo2018,Stvilia2007,Tahar2023,Kahn2016} \\
\cdashline{2-4}[0.5pt/2.5pt]
\ & compliance & \footnotesize The extent to which the dataset and data values are in accordance with laws, regulations or standards. (From \cite{dama2020}) &\footnotesize \cite{iso2500025012,Syed2023,gudivada2017data,Johnson2015,Kim2003,Chen2014,Schmidt2021,Eder2020,Stvilia2007,Loshin2011} \\
\cdashline{1-4}
\raggedright logical consistency & \ & \footnotesize The extent to which data values are plausible and records are semantically consistent. &\footnotesize \cite{Batini2015} \\
\cdashline{2-4}[0.5pt/2.5pt]
\ & semantic consistency & \footnotesize The extent to which records are free of contradictions. &\footnotesize \cite{Qi2021,Cai2015,Syed2023,Johnson2015,Batini2015,Schmidt2021,Mashoufi2023,Chen2014,Vanbrabant2019,iso2500025012,Ramasamy2020,Bloland2019,Kim2003,Redman1996,Eder2020,Loshin2011,Sidi2012,chan2010electronic,Bian2020,Lewis2023,gudivada2017data,Juddoo2018,Stvilia2007,Tahar2023,Weiskopf2013} \\
\cdashline{2-4}[0.5pt/2.5pt]
\ & plausibility & \footnotesize The extent to which data values match knowledge of the real world. (From  \cite{dama2020}) &\footnotesize \cite{Lewis2023,Sidi2012,Syed2023,Wang1996,Batini2015,Juddoo2018,Pipino2002,Tahar2023,Loshin2011,Kahn2016,Bian2020,Weiskopf2013} \\
\cdashline{1-4}
\raggedright distribution consistency & \ & \footnotesize The extent to which distributions are stable. &\footnotesize \\
\cdashline{2-4}[0.5pt/2.5pt]
\ & homogeneity & \footnotesize The extent to which distributions are stable among relevant subsets. &\footnotesize \cite{Pan2023,Redman1996,Syed2023} \\
\cdashline{2-4}[0.5pt/2.5pt]
\ & distribution drift & \footnotesize The extent to which new data matches the distribution of existing data. &\footnotesize \cite{sebastian2012measuring,Syed2023,Liu2023,ovadia2019can,Loshin2011,Sidi2012,Bian2020} \\
\hline
\end{tabular}
\caption{Details for the cluster \textit{Consistency} of the METRIC-framework. The hierarchical order of all dimensions and subdimensions included in the cluster is displayed in the first two columns. The third column provides a definition for each of these terms. The fourth column provides all references of the literature corpus connected to the corresponding term. A provided reference indicates that the dimension, subdimension or a different term with the same intended meaning (see \nameref{sec:dataextraction} for clarification) is discussed in that cited article. References on the cluster level indicate that the term is used as a broader concept.}
\label{tab:dqd_to_papers5}
\end{table}
\clearpage
\begin{table}[ht!]
\begin{tabular}{p{10.7cm} p{5.3cm} p{3.7cm} p{3.5cm}}
\bf Data Management& \ & \ &\footnotesize \\
\end{tabular}
\begin{tabular}{| p{3.7cm} | p{3.3cm} | p{3.7cm} |  p{3.5cm} |}
\hline
Dimension & Subdimension & Definition & References \\
\hline
\raggedright documentation & \ & \footnotesize Metadata about the dataset. &\footnotesize \cite{Cai2015,Syed2023,gudivada2017data,Johnson2015,Batini2015,Eder2020,Redman1996,Loshin2011,Sidi2012} \\
\cdashline{1-4}
\raggedright security & \ & \footnotesize A characteristic that ensures that unauthorized changes to the dataset are prohibited. &\footnotesize \cite{Sidi2012,Batini2015,Gao2016,Pipino2002,Chen2014,Stvilia2007,Wang1996} \\
\cdashline{1-4}
\raggedright privacy & \ & \footnotesize  A characteristic that ensures that the dataset is recorded and stored in a way that protects personal data. &\footnotesize \cite{iso2500025012,gudivada2017data,Gao2016,Ramasamy2020,Chen2014,Loshin2011} \\ 
\cdashline{1-4}
\raggedright FAIR principles & \ & \footnotesize The FAIR Principles \cite{Wilkinson2016} require a dataset to be findable, accessible, interoperable and reusable. &\footnotesize \cite{Cai2015,Syed2023,Batini2015,Gao2016,Chen2014,Mashoufi2023,yoon2000managing,nahm2012data,iso2500025012,Liu2023,Ramasamy2020,Pipino2002,Redman1996,Loshin2011,Sidi2012,Wang1996,gudivada2017data,Juddoo2018,Stvilia2007} \\
\hline
\end{tabular}
\caption{Details for \textit{Data Management} as a prerequisite for the METRIC-framework. All dimensions included under data management are displayed in the first column. The third column provides a definition for each of these terms. The fourth column provides all references of the literature corpus connected to the corresponding term. A provided reference indicates that the dimension or a different term with the same intended meaning (see \nameref{sec:dataextraction} for clarification) is discussed in that cited article.}
\label{tab:dqd_to_papers6}
\end{table}

%% file: sections/section6others.tex
\section*{}\label{sec:others}

\subsection*{Acknowledgements}
We would like to thank Stefan Haufe for valuable input on the manuscript. We further thank the project partners of the TEF-Health project for feedback on our study. 

\subsection*{Author contributions}
D.S. and T.S. designed and supervised the study. D.S., K.B., M.S., A.K. carried out the theoretical methods and analysed the data. M.S., A.K. extracted the data. D.S., K.B., M.S., A.K., T.S. wrote the manuscript.

\subsection*{Competing interests}
The authors declare no competing interests.

\subsection*{Funding}
The authors acknowledge the funding by the EU TEF-Health project which is part of the Digital Europe Programme of the EU (DIGITAL-2022-CLOUD-AI-02-TEF-HEALTH).

\subsection*{Additional information}
The extracted data and terms from the literature corpus that served as a basis for the METRIC-framework are provided in a supplementary Excel file.